\documentclass[sigconf]{acmart}

\usepackage{subfig}

\AtBeginDocument{%
  }

\setcopyright{acmlicensed}
\copyrightyear{2026}
\acmYear{2026}
\acmDOI{XXXXXXX.XXXXXXX}

\acmConference[WSDM'26]{Make sure to enter the correct
  conference title from your rights confirmation email}{February 22--26,
  2026}{Boise, Idaho, USA}

\acmISBN{978-1-4503-XXXX-X/2018/06}

\begin{document}

\title{SARC: Sentiment-Augmented Deep Role Clustering for Fake News Detection}

\author{Jingqing Wang}
\affiliation{%
  \institution{College of Computer Science, Chongqing University}
  \city{Chongqing}
  \country{China}
}
\email{wangjq972@gmail.com}

\author{Jiaxing Shang}
\authornotemark[1]
\affiliation{%
  \institution{Department of Computer Science, University of Exeter}
  \city{Exeter}
  \country{UK}
}
\email{j.shang@exeter.ac.uk}

\author{Rong Xu}
\affiliation{
  \institution{College of Computer Science, Chongqing University}
  \city{Chongqing}
  \country{China}
}
\email{rongxu126@gmail.com}

\author{Fei Hao}
\affiliation{
 \institution{School of Artificial Intelligence and Computer Science, Shaanxi Normal University}
 \city{Xi'an}
 \country{China}
}
\email{feehao@gmail.com}

\author{Tianjin Huang}
\affiliation{
  \institution{Department of Computer Science, University of Exeter}
  \city{Exeter}
  \country{UK}
}
\email{t.huang2@exeter.ac.uk}

\author{Geyong Min}
\affiliation{
  \institution{Department of Computer Science, University of Exeter}
  \city{Exeter}
  \country{UK}}
\email{g.min@exeter.ac.uk}

\renewcommand{\shortauthors}{Wang et al.}

\begin{abstract}
Fake news detection has been a long-standing research focus in social networks. Recent studies suggest that incorporating sentiment information from both news content and user comments can enhance detection performance. However, existing approaches typically treat sentiment features as auxiliary signals, overlooking role differentiation, that is, the same sentiment polarity may originate from users with distinct roles, thereby limiting their ability to capture nuanced patterns for effective detection. To address this issue, we propose \textbf{SARC}, a \textbf{S}entiment-\textbf{A}ugmented \textbf{R}ole \textbf{C}lustering framework which utilizes sentiment-enhanced deep clustering to identify user roles for improved fake news detection. The framework first generates user features through joint comment text representation (with BiGRU and Attention mechanism) and sentiment encoding. It then constructs a differentiable deep clustering module to automatically categorize user roles. Finally, unlike existing approaches which take fake news label as the unique supervision signal, we propose a joint optimization objective integrating role clustering and fake news detection to further improve the model performance. Experimental results on two benchmark datasets, RumourEval-19 and Weibo-comp, demonstrate that SARC achieves superior performance across all metrics compared to baseline models. The code is available at: https://github.com/jxshang/SARC.
\end{abstract}



\maketitle

\section{Introduction}
The rapid advancement of the Internet and mobile communication technologies has fundamentally reshaped human communication patterns, positioning social media platforms as the dominant channels for information exchange \cite{notarmuzi2022universality,shang2025dvcae,xiong2025sdvd}. By removing the temporal and spatial constraints of information dissemination, these platforms enable instant communication across global audiences, fostering positive societal impacts such as advancing social marketing \cite{bhimaraju2024fractional,chakraborty2023learning} and promoting social equality \cite{akpuokwe2024innovating}. However, the same mechanisms that facilitate rapid and wide-reaching communication also create fertile ground for the proliferation of fake news, which can undermine public trust and distort social discourse \cite{bovet2019influence,olan2024fake}. With the emergence of large language models (LLMs), the challenge has intensified, as fake news content has become increasingly sophisticated and harder to detect \cite{wu2024fake,hu2025llm}. Consequently, fake news detection on social networks has received growing attention in recent years. Addressing this challenge demands effective detection methods capable of operating in the dynamic and complex environment of online interactions. Most state-of-the-art approaches are based on deep learning technologies and can be broadly categorized into content-based, social-context-based, and LLM-based methods. For instance, Hu et al. \cite{hu2021compare} constructed a topic-entity-sentence heterogeneous graph enhanced with knowledge graphs to reveal differences between true and fake news. Cui et al. \cite{cui2024propagation} proposed RAGCL, an adaptive graph contrastive learning model that captures structural features of rumor propagation trees through adaptive augmentation and contrastive learning. Zhou et al. \cite{zhou2025collaborative} integrated LLM-based retrieval of up-to-date knowledge with small language models for collaborative fake news detection.

\begin{figure*}[htbp]
    \centering
    \includegraphics[width=0.8\linewidth]{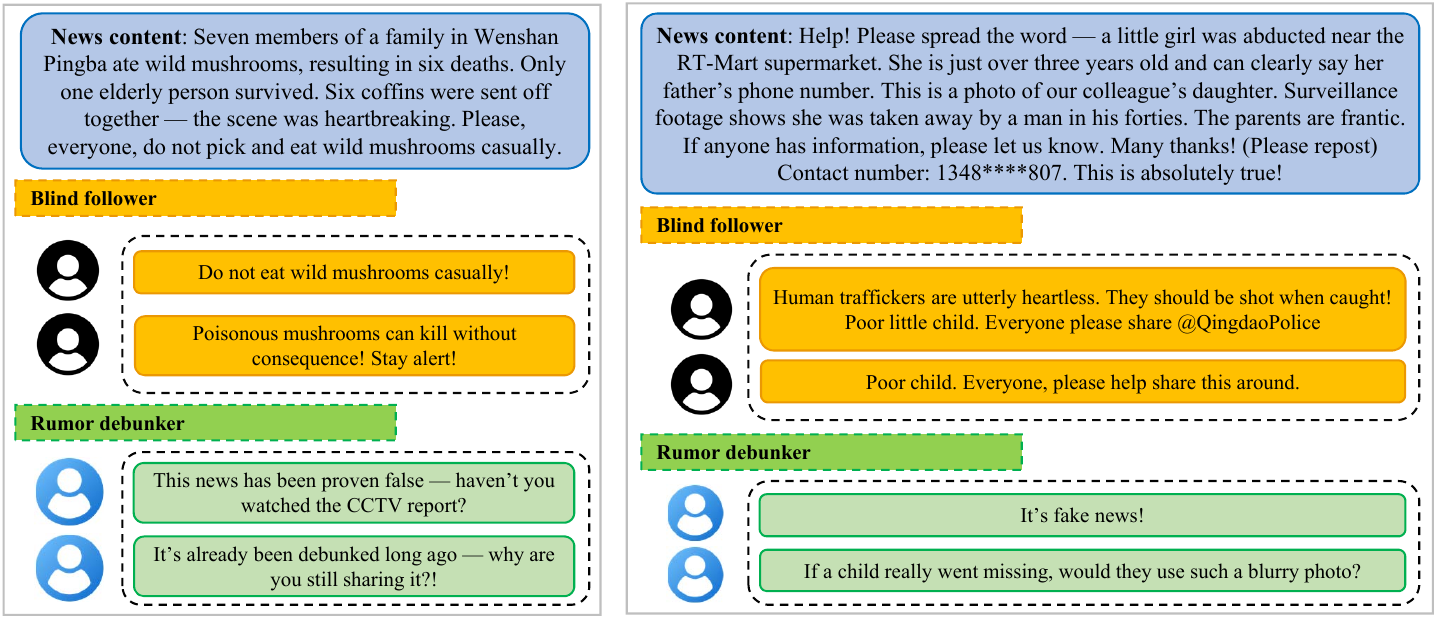}
    \caption{Motivational example showing two pieces of fake news and user comments from the Weibo-comp dataset.}
    \Description{Motivational example.}
    \label{fig:motivation}
\end{figure*}

\textbf{\emph{Motivation.}} Despite notable progress achieved by deep learning-based fake news detection models, most existing studies still focus primarily on the semantic features of news content or shallow interaction signals between news content and user comments. Recent research has demonstrated that incorporating sentiment information from news content and user comments can provide valuable clues for identifying news authenticity \cite{shu2019defend,zhou2023does,zhu2024exploring}. For instance, Zhang et al. \cite{zhang2021mining} extracted sentiment features representing dual emotions between news content and user comments to improve fake news detection. Jia et al. \cite{jiang2024makes} jointly modeled sentiment and stance features, proposing a multi-task learning framework that enhances detection performance by training additional sentiment and stance classifiers. Zhang et al. \cite{zhang2023sentence} introduced a graph attention network-based model to capture mutual influences of sentiment between sentences within long-form news for fake news detection. However, these sentiment-based approaches typically treat sentiment and stance information as auxiliary features, overlooking the critical fact of \emph{role differentiation}, that is, the same sentiment polarity may originate from users with distinct roles. As a motivational example, Figure \ref{fig:motivation} illustrates two fake news cases and their associated user comments from the Weibo-comp dataset, where comments expressing similar sentiment polarity (negative) stem from different user roles. Specifically, ``blind followers'' express negative emotions (e.g., anger) toward the news content without questioning its authenticity, whereas ``rumor debunkers'', despite sharing a similar negative sentiment, explicitly express doubt about the news's veracity. Neglecting such nuanced patterns can substantially limit the ability of existing models to detect fake news effectively.

\textbf{\emph{Challenge.}} To incorporate the roles behind user comments into fake news detection, two critical challenges must be addressed. \emph{Challenge 1: how to identify user roles when ground-truth annotations are unavailable?} To the best of our knowledge, none of the existing fake news datasets provide role labels for user comments. While stance detection methods \cite{jiang2024makes,yang2025llm} implicitly categorize user comments into predefined roles (e.g., supporter or opponent), such roles are hard-coded and cannot be flexibly adapted to specific tasks or datasets. \emph{Challenge 2: how to effectively integrate the role identification task into the fake news detection pipeline to improve overall performance?} The key to addressing this challenge lies in designing an effective framework that jointly optimizes the role identification and fake news detection tasks.

\textbf{\emph{Solution.}} To address the above challenges, we propose \textbf{SARC}, a \textbf{S}entiment-\textbf{A}ugmented deep \textbf{R}ole \textbf{C}lustering framework that automatically identifies  user roles in news comment sections to enhance fake news detection. Specifically, SARC incorporates an unsupervised, learnable deep role clustering module that jointly leverages sentiment features and textual content of comments to model collective behavioral patterns of users. By adopting an unsupervised deep clustering approach, SARC overcomes the limitations of traditional supervised role classification methods, which heavily rely on annotated datasets and suffer from label scarcity. Moreover, SARC introduces an integrated loss function that jointly optimizes the clustering loss and the fake news detection loss, thereby tightly coupling role identification with fake news detection and improving both accuracy and generalization. The SARC framework consists of four core components: (1) Initial Feature Representation Module, which establishes semantic foundations for news text and user comments using pretrained word embeddings while deriving sentiment polarity features for individual users; (2) Text Encoding Module, which utilizes dual-channel BiGRU networks with self-attention to extract deep semantic features from news content and temporal patterns from comments; (3) Dynamic Role Clustering Module, which creatively integrates differentiable deep clustering algorithms to automatically partition sentiment-augmented comment features into latent user behavioral patterns, while adaptively optimizing cluster distributions; (4)  News Classification Module, which fuses news semantics with role features via feature concatenation and applies multilayer perceptrons for final classification. 

The main contributions of this paper are as follows:

\begin{itemize}
\item We provide a novel perspective by modeling collective cognition in news comment sections through user roles. Unlike prior works that treat comments merely as auxiliary textual features, SARC conceptualizes role differentiation as dynamic representations of group cognition, quantifying user affiliations (i.e., roles) via learnable role prototype vectors (i.e., cluster centers).

\item We propose a differentiable deep clustering algorithm for the unsupervised categorization of user roles. In contrast to existing methods that rely solely on fake news labels for supervision, SARC introduces a joint optimization objective that integrates role clustering with fake news detection. By jointly optimizing clustering and detection losses, SARC tightly couples role identification with detection, thereby enhancing both accuracy and generalization.

\item We conduct comprehensive experiments on two real-world datasets, comparing SARC with state-of-the-art methods. Results demonstrate that SARC achieves substantial improvements, ranging from 4.9\% to 15.3\% in detection accuracy over competitive baselines. Ablation studies further confirm the effectiveness of the proposed modules.
\end{itemize}

\section{Ralated Work}
\subsection{Fake News Detection}
In recent years, fake news detection on social media has garnered significant attention. Based on the types of information utilized by detection models, existing research can be broadly categorized into three main approaches: content-based detection, social-context-based detection, and LLM-based detection.

Content-based approaches focus on analyzing various news content, such as headlines, body text, images, videos, and audio. Language-based methods use deep neural networks to extract linguistic features that distinguish true and fake news through stylistic disparities \cite{potthast2017stylometric}, though such boundaries are increasingly blurred by advanced editorial practices and LLM/AIGC technologies. Some studies enhance linguistic feature via multi-task learning, such as MTEFN \cite{choudhry2022emotion}, which jointly optimizes emotion classification and fake news detection by leveraging emotion-related cues. The second category comprises knowledge-enhanced methods that integrate external knowledge related to news content. For instance, Hu et al. \cite{hu2021compare} built a topic-entity-sentence heterogeneous graph using knowledge graphs, Yang et al. \cite{yang2025macro} applied cross-domain transfer to address domain shift, Wang et al. \cite{wang2025collaboration} fused generated and original comments to implicitly incorporate expert knowledge. Multi-modal fusion methods combine textual, visual, and auditory data, evolving from early text–image combinations \cite{meel2021han,zhang2024reinforced,simonyan2014very} to advanced mechanisms such as residual-aware compensation \cite{yu2025racmc}, hyperbolic representation framework \cite{feng2025mhr}, multi-expert optimization \cite{shen2025gamed}, multi-granularity clue alignment \cite{guo2025each}, and interactive gating for modality interaction \cite{liu2025modality}.

Social-context-based detection techniques include stance-based and propagation-based methods. Stance-based approaches model user reactions, such as attitudes, skepticism, or positional tendencies, to aid detection. For example, Ayoobi et al. \cite{ayoobi2024seeing} proposed ESAS, a metric quantifying user skepticism toward LLM-generated fake news, while HSA-BLSTM \cite{guo2018rumor} uses a hierarchical attention network to extract stance features from posts and sub-events within social contexts. Propagation-based methods instead model dissemination patterns, such as structures, dynamics, or network topologies. Xu et al. \cite{xu2024harnessing} proposed NAGASIL which simulates fake news spread to identify optimal debunkers. Cui et al. \cite{cui2024propagation} proposed RAGCL which encodes rumor propagation tree features via adaptive graph contrastive learning. Zhang et al. \cite{zhang2024bayesian} introduced BLC to handle uncertainty and noisy relations, while Kim et al. \cite{kim2025revisiting} proposed DAWN, which weights edges using engagement timeliness to refine propagation modeling. Sun et al. \cite{sun2025unifying} addressed social context confounders through adversarial multi-debiasing, and Chen et al. \cite{chen2025birds} developed ReTIP, retrieving relevant participants and their interaction contexts to capture the role of social actors in dissemination.

With the rise of large language models (LLMs), LLM-based fake-news detection has become a new paradigm. They leverage pre-trained knowledge for deep semantic understanding and factual verification, even with limited data. From the perspective of data augmentation and style adaptation, Park et al. \cite{park2025adversarial} proposed adversarial style augmentation to generate diverse stylistic variants of fake news, while Nan et al. \cite{nan2024let} generated simulated user comments to supplement scarce early-stage data. For explainability, Wang et al. \cite{wang2024explainable} used LLMs to produce human-readable reasoning chains, and Liu et al. \cite{liu2025truth} introduced TruEDebate, an LLM-driven multi-agent debate framework enhancing both explainability and detection performance. In knowledge retrieval and generative assistance, Zhou et al. \cite{zhou2025collaborative} combined LLMs for retrieving up-to-date knowledge with small language models for collaborative detection, Zhang et al. \cite{zhang2025llms} analyzed LLM hallucination patterns to identify LLM-generated fake news, and Hu et al. \cite{hu2024bad} proposed ARG which employs dual cross-attention to fuse news content with multi-perspective analyses generated by LLMs.

\subsection{Deep Clustering}
Clustering, as a typical unsupervised learning technique, aims to partition data samples into groups (clusters) based on similarity metrics. However, traditional clustering methods struggle with high-dimensional data, prompting the development of deep clustering approaches.

For instance, McConville et al. \cite{mcconville2021n2d} proposed a method combining autoencoder embeddings with local manifold learning. This approach preserves autoencoder representational capacity while leveraging the UMAP algorithm \cite{mcinnes2018umap} to uncover local geometric structures in latent spaces, followed by Gaussian mixture models for clustering. Van Gansbeke et al. \cite{van2020scan} proposed SCAN which employs a two-stage framework for unsupervised image classification: It first learns semantic-invariant features via self-supervised learning (e.g., instance discrimination), then optimizes cluster assignments through nearest-neighbor prior-based clustering loss with consistency constraints and entropy regularization. Liu et al. \cite{liu2024end} proposed ELCRec which addresses scalability in recommendation systems by integrating user behavior representation learning with end-to-end learnable clustering. This method initializes cluster centers as learnable neurons, designs learnable clustering loss functions to disentangle user intents via cluster separation, and enhances representation-clustering synergy through intent-aware contrastive learning.

\begin{figure*}[htbp]
    \centering
    \includegraphics[width=\linewidth]{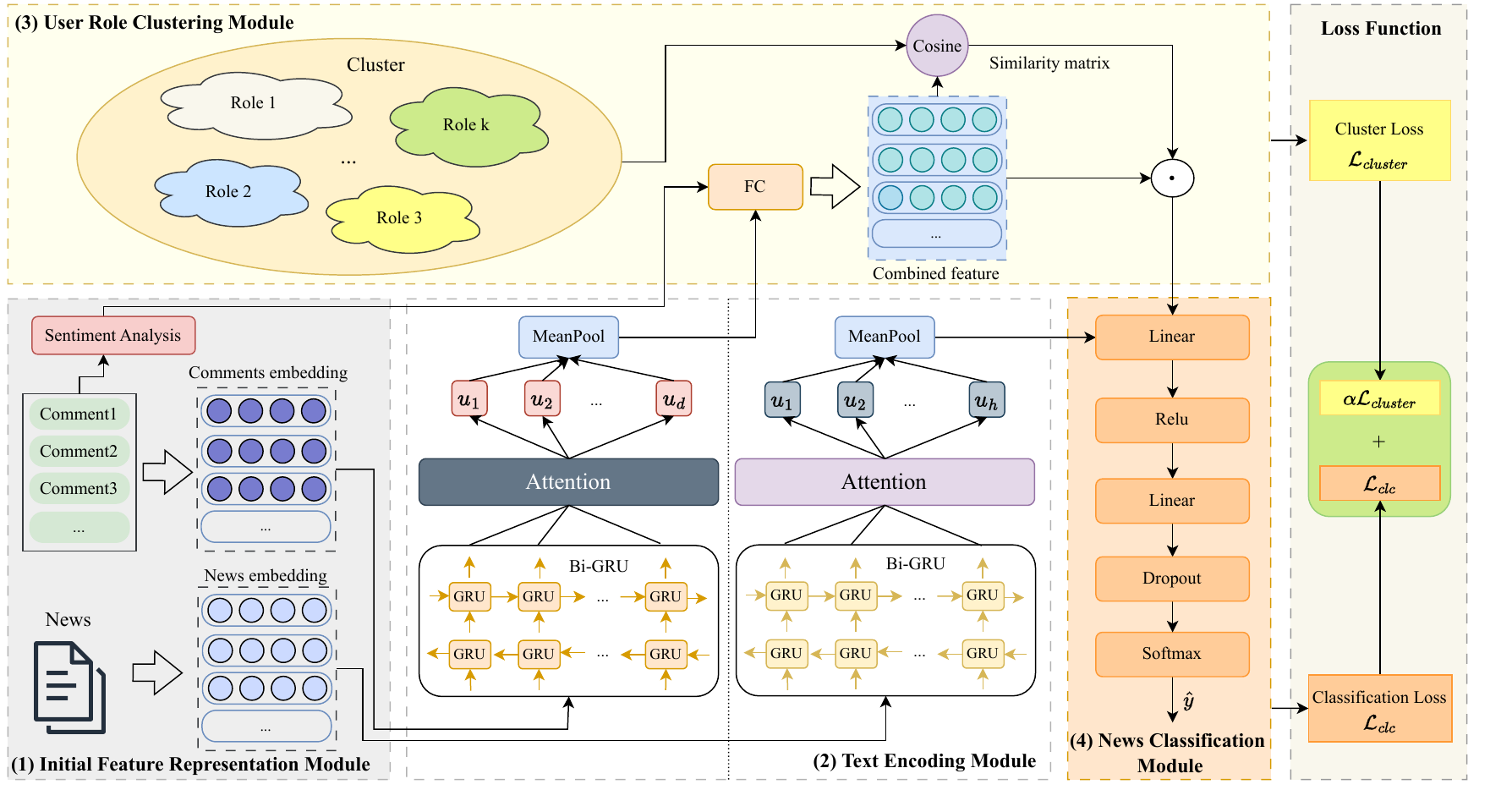}
    \caption{The proposed SARC framework which consists of four components: (1) Initial Feature Representation Module, (2) Text encoding module, (3) User Role Clustering Module, (4) News Classification Module.}
    \Description{The proposed SARC framework which consists of four components: (1) Initial Feature Representation Module, (2) Text encoding module, (3) User Role Clustering Module, (4) News Classification Module.}
    \label{model_struct}
\end{figure*}

\section{Problem Statement}
Given a news article $ A = \{w_i\}_{i=1}^{N} $ (containing a word sequence of length $ N $) and its associated comment set $ C = \{c_j\}_{j=1}^T $ (with $ T $ user comments), where each comment $ c_j $ is represented as $ c_j = \{w_l^j\}_{l=1}^{L_j} $ with $ w_l^j $ denoting the $l$-th word in the $j$-th comment and $L_j$ denoting the comment length, the objective of fake news detection is to predict the veracity label $ \hat{y} = f(A, C; \Theta) $ through a detection model $ f(\cdot) $ parameterized by $ \Theta $.

\section{Methodology}
In this section, we will introduce in detail the proposed SARC framework, as shown in Figure \ref{model_struct}, which consists of four logically connected components: (1) the Initial Feature Representation Module; (2) the Text Encoding Module; (3) the User Role Clustering Module; and (4) the News Classification Module.

First, SARC employs the Initial Feature Representation Module to generate the feature representations of the news content and user comments, obtaining the sentiment polarity representation of the comments. Subsequently, the Text Encoding Module is used to extract the deep semantic features of the news and the sequential features of the comments respectively. Then, the comment features are enhanced with emotional features. Furthermore, the User Role Clustering Module is adopted to dynamically divide user roles for the sentiment-enhanced comment features in the latent space. Finally, the news semantic vectors and the user role clustering features are fused, and the classification of fake news is achieved through the News Classification Module.

\subsection{Initial Feature Representation Module}
The initial feature representation module processes news articles and user comments to obtain corresponding embeddings and sentiment polarity representations.

\subsubsection{News Embedding}
The objective is to convert news text from natural language into vector representations. As news articles typically consist of long texts containing event details such as time, location, and causality, we initialize the embedding layer using word vectors pretrained on large corpora to better capture semantic information. For a news article $A = \{w_i\}_{i=1}^N$ composed of $N$ words, the embedding layer initialized with pretrained word vectors generates word embeddings $\mathbf{T_A} = \{\mathbf{t_i}\}_{i=1}^N$, where $\mathbf{t_i}$ represents the word vector corresponding to $w_i$.

\subsubsection{Comment Embedding}
Sharing the same objective of converting text into machine-understandable vector forms, comment embeddings employ separate embedding layers from news embeddings. This design addresses the divergent focuses between news content (emphasizing factual elements like institutional names, locations, and temporal markers) and user comments (dominated by opinions, emotions, and internet slang). For the comment set $C = \{c_j\}_{j=1}^T$ associated with the news, each comment $c_j$ is encoded through a pretrained embedding layer as $\mathbf{T_{c_j}} = \{\mathbf{t_{i}^j}\}_{i=1}^{L_{c_j}}$, where $\mathbf{t_{i}^j}$ denotes the word vector corresponding to $w_i^j$.

\subsubsection{Sentiment Analysis}
As emotions drive human behavior and fake news demonstrates higher emotional intensity, stronger negativity, and weaker positivity compared to true news \cite{zhou2023does}, user comments often reflect emotional responses \cite{horner2023emotions}. We employ a pretrained sentiment analysis model to assign ternary sentiment polarity scores (negative/neutral/positive) to comments. The comment set $C = \{c_j\}_{j=1}^T$ is thereby transformed into sentiment polarity representations $E = \{e_j\}_{j=1}^T$, where $e_j$ corresponds to the sentiment polarity of comment $c_j$.

\subsection{Text Encoding Module}
Fake news is deliberately crafted to disseminate false information, often employing sensational language to attract attention and generate traffic, resulting in distinct stylistic differences from true news \cite{potthast2017stylometric}. To capture these disparities, we employ two text encoders to learn semantic representations for news content and user comments respectively. 

We utilize Gated Recurrent Units (GRU) \cite{cho2014learning} to model long-range dependencies and maintain persistent memory. Bidirectional GRU \cite{chung2014empirical} further enhances context awareness by encoding text sequences from both directions.

Given word embeddings $\mathbf{T} = \{\mathbf{t_i}\}_{i=1}^L$ for a text sequence (news or comment) of length $L$, the forward GRU ($\overrightarrow{GRU}$) and backward GRU ($\overleftarrow{GRU}$) compute hidden states as:

\begin{equation}\label{BiGRU}
    \begin{aligned}
        \overrightarrow{\mathbf{h}}_i &= \overrightarrow{\text{GRU}}    (\mathbf{t}_i, \overrightarrow{\mathbf{h}}_{i-1}), \quad \forall i \in  [1, L] \\
        \overleftarrow{\mathbf{h}}_i &= \overleftarrow{\text{GRU}}(\mathbf{t}_i,    \overleftarrow{\mathbf{h}}_{i+1}), \quad \forall i \in [L, 1]
    \end{aligned}
\end{equation}

The bidirectional context-aware representation is obtained through concatenation: $\mathbf{h}_i = [\overrightarrow{\mathbf{h}}_i, \overleftarrow{\mathbf{h}}_i]$, which encapsulates full-sentence semantics centered at word $w_i$. To amplify intentionally crafted words (e.g., exaggerations) in fake news, we introduce self-attention \cite{vaswani2017attention}. Given GRU hidden states $\mathbf{H} = [\mathbf{h}_1, \cdots, \mathbf{h}_L] \in \mathbb{R}^{L \times 2d_h}$ ($d_h$: GRU hidden dimension), the attention matrix $\mathbf{A} \in \mathbb{R}^{L \times L}$ is computed via scaled dot-product:
\begin{equation}\label{selfAttention}
    \mathbf{A}_{ij} = \frac{\exp(\mathbf{q}_i^\top\mathbf{k}_j/\sqrt{d_k})} {\sum_{k=1}^L\exp(\mathbf{q}_i^\top\mathbf{k}_k/\sqrt{d_k})}, \quad
    \begin{cases}
        \mathbf{q}_i = \mathbf{W}_Q\mathbf{h}_i \\
        \mathbf{k}_j = \mathbf{W}_K\mathbf{h}_j
    \end{cases}
\end{equation}
where $\mathbf{W}_Q, \mathbf{W}_K \in \mathbb{R}^{2d_h \times d_k}$ are learnable projection matrices. The enhanced representation combines attention weights with residual connections:
\begin{equation}\label{SA_out}
    \mathbf{u}_i = \sum_{j=1}^L \mathbf{A}_{ij}\mathbf{h}_j + \mathbf{h}_i
\end{equation}

Mean pooling aggregates sequence-level features into fixed-dimensional vectors:
\begin{equation}\label{pooling}
    \mathbf{v} = \frac{1}{L}\sum_{i=1}^L \mathbf{u}_i, \quad \mathbf{v} \in     \mathbb{R}^{2d_h}
\end{equation}

Given the semantic divergence between news content and user comments, we implement dual encoders with shared architecture but independent parameters:
\begin{equation}\label{eq:dual_encoders}
    \begin{aligned}
        \mathbf{v}_A &= \text{encode}_A(\mathbf{T}_A) \\ &= \text{MeanPool} (\text{SelfAttn}(\text{BiGRU}(\mathbf{T}_A))) \\
        \mathbf{V}_C &= [\mathbf{v}_{c_1}, ..., \mathbf{v}_{c_T}] \\&=     [\text{encode}_C(\mathbf{T}_{c_1}), ...,    \text{encode}_C(\mathbf{T}_{c_T})]
    \end{aligned}
\end{equation}

\subsection{Dynamic Role Clustering Module}
During fake news dissemination, user comments often exhibit role differentiation (e.g., spammers, skeptics, blind followers), which differs from true news propagation. To exploit this phenomenon, we employ end-to-end deep clustering algorithms for user role categorization, addressing the limitations of traditional clustering methods in high-dimensional data processing.

Given user comment features $\mathbf{V}_C$ and sentiment features $E$, we first concatenate the two features:
\begin{equation}
    \begin{aligned}
        \mathbf{V}_C^\text{e} &= [\mathbf{v}_{c_1}^\text{e},    \mathbf{v}_{c_2}^\text{e}, ..., \mathbf{v}_{c_T}^\text{e}]\\   
        \mathbf{v}_{c_j}^{\text{e}} &= \big[ \mathbf{v}_{c_j},\, e_j \big]
    \end{aligned}
\end{equation}
where $\mathbf{v}_{c_j} \in \mathbf{V}_C,\; e_j \in E$. 

These features are projected into clustering space via fully connected layer:
\begin{equation}\label{eq:cluster_proj}
    \mathbf{X} = \text{FC}(\mathbf{V}_C^\text{e})
\end{equation}
where $\mathbf{X}\in \mathbb{R}^{T \times d_p}$, FC denotes a fully connected layer, $T$ denotes comment count and $d_p$ denotes the clustering space dimension.

We define $K$ trainable cluster centers $\mathbf{M} = [\mu_1, \mu_2, ..., \mu_K] \in \mathbb{R}^{K \times d_p}$ and apply L2 normalization:
\begin{equation}
    \begin{aligned}
        \mathbf{\tilde{X}} &= \text{L2Norm}(\mathbf{X}, \text{dim}=-1) \\
        \mathbf{\tilde{M}} &= \text{L2Norm}(\mathbf{M}, \text{dim}=-1)
    \end{aligned}
\end{equation}

We use cosine similarity as the distance between the feature and the cluster center. The cosine similarity matrix $\mathbf{S} = \big[s_{jk}\big]_{1 \leq j \leq T}^{1 \leq k \leq K } \in \mathbb{R}^{T \times K}$ is computed as:
\begin{equation}\label{eq:cos_sim}
    s_{jk} = \mathbf{\tilde{x}}_j^\top \mathbf{\tilde{\mu}}_k \in [-1, 1]
\end{equation}
Where $\tilde{x}_j $ and ${\tilde{\mu}}_k$ denote the $j\textbf{-th}$ row of $\mathbf{\tilde{X}}$ and the $k\textbf{-th}$ row of $\mathbf{\tilde{M}}$.

Temperature-scaled soft assignment probabilities are computed as:
\begin{equation}\label{eq:soft_assign}
    q_{jk} = \frac{\exp(s_{jk} \times \tau)}{\sum_{k'=1}^K \exp(s_{jk'} \times  \tau)}
\end{equation}
where $\tau$ is a learned temperature coefficient. This yields the soft assignment matrix $\mathbf{Q}= \big[ q_{jk} \big]_{ 1 \leq j \leq T}^{1 \leq k \leq K } \in \mathbb{R}^{T \times K}$.

Cluster-level semantic representations are aggregated through weighted summation:
\begin{equation}\label{eq:cluster_agg}
    \begin{aligned}
        \mathbf{C}_K &= \mathbf{Q}^\top \mathbf{\tilde{X}} \\
    \end{aligned}
\end{equation}
where $\mathbf{C}_K \in \mathbb{R}^{K \times d_p}$ contains global semantic features for each role category.

\subsection{News Classification Module}
This module detects fake news by combining the semantic features of news content with user role features. The two feature sets are merged using a linear transformation and then passed to a classifier.

Given the news semantic feature vector $\mathbf{v}_A \in \mathbb{R}^{2d_A}$ from Eq. \eqref{eq:dual_encoders} and user role cluster feature matrix $\mathbf{C}_K \in \mathbb{R}^{K \times d_p}$ from Eq. \eqref{eq:cluster_agg}, fusing news semantic features and user role features through flattening and concatenation:
\begin{equation}\label{eq:feature_fusion}
    \mathbf{f} = [\mathbf{v}_A; \text{vec}(\mathbf{C}_K)] \in \mathbb{R}^{2d_A + K  \cdot d_p}
\end{equation}

The classification decision is realized through deep nonlinear mapping:
\begin{equation}\label{eq:classifier}
    \begin{aligned}
        \mathbf{h}_1 &= \text{Dropout}(\sigma(\mathbf{W}_1 \mathbf{f} +     \mathbf{b}_1))  \\
        \hat{y} &= \text{Softmax}(\mathbf{W}_2 \mathbf{h}_1 + \mathbf{b}_2)
    \end{aligned}
\end{equation}
where $\sigma$ denotes the ReLU activation function.

\subsection{Loss Function}
The model achieves collaborative learning of fake news detection and user role cluster through multi-task joint optimization. The total loss combines supervised classification loss and unsupervised clustering loss:
\begin{equation}
    \mathcal{L} = \mathcal{L}_{\text{cls}} + \alpha\mathcal{L}_{\text{cluster}}
\end{equation}
where $\alpha$ balances the importance of unsupervised clustering loss.

\subsubsection{Classification Loss}
We employ L2-regularized cross-entropy loss for each sample:
\begin{equation}\label{eq:cls_loss}
    \mathcal{L}_{\text{cls}} = -\frac{1}    {B}\sum_{b=1}^B\sum_{c=1}^My_{bc}\log(p_{bc})
\end{equation}
where $y_{bc}$ denotes the true class label and $p_{bc}$ is the predicted  probability for class $c$.

\subsubsection{Unsupervised Clustering Loss}
The clustering loss consists of intra-class loss $\mathcal{L}_{\text{intra}}$ and inter-class loss $\mathcal{L}_{\text{inter}}$:
\begin{equation}
    \mathcal{L}_{\text{cluster}} = \mathcal{L}_{\text{intra}} +     \mathcal{L}_{\text{inter}}
\end{equation}

To prevent early-stage errors from being reinforced \cite{liu2024end}, the  intra-class loss is designed to pull features toward all cluster centers:
\begin{equation}\label{eq:intra_loss}
    \mathcal{L}_{\text{intra}} = -\frac{1}{B \cdot T} \sum_{b=1}^B \sum_{j=1}^{T}   \sum_{k=1}^K q_{bjk} \cdot \underbrace{\mathbf{\tilde{x}}_{bj}^\top   \mathbf{\tilde{\mu}}_k}_{\text{cosine similarity}}
\end{equation}

The inter-class loss separates cluster centers:
\begin{equation}\label{eq:inter_loss}
    \mathcal{L}_{\text{inter}} = \frac{1}{K(K-1)} \sum_{k=1}^K \sum_{\substack{k'=1     \\ k' \neq k}}^K {\mathbf{\tilde{\mu}}_k^\top \mathbf{\tilde{\mu}}_{k'}}
\end{equation}

Here $q_{bjk} \in [0,1]$ represents the probability of comment $j$ in batch $b$ being assigned to cluster $k$, with $\mathbf{\tilde{x}}_{bj} =  \frac{\mathbf{x}_{bj}}{\|\mathbf{x}_{bj}\|_2}$ and $\mathbf{\tilde{\mu}}_k = \frac{\mu_k}{\|\mu_k\|_2}$ being L2-normalized features and cluster centers respectively.

\section{Experimental Evaluation}
\subsection{Datasets}
To evaluate the performance of SARC in fake news detection, we conduct experiments on the Chinese dataset Weibo-comp \cite{DataFountain2020COVID} and the English dataset RumourEval-19 \cite{gorrell2019determining}. 

\subsubsection{Weibo-comp} 
It is a multimodal dataset containing Weibo posts with original texts, comments, images, and domain categories. It covers news content from eight domains (e.g., technology, politics, military, etc.) on Weibo platform, including COVID-19-related information. The dataset contains three categories: real, fake, and undecidable news. Here, we use real and fake news with corresponding comments for experiments.

\subsubsection{RumourEval-19} 
It originates from the SemEval-2019 rumor detection task, containing news posts from Twitter and Reddit. It categorizes news into three classes: real, fake, and unverified. Detailed dataset statistics are shown in Table \ref{tab:dataset_stats}.       \begin{table}[htbp]
    \centering
    \vspace{10pt}
    \begin{tabular}{l|cc}
        \hline
        \multicolumn{1}{c|}{\textbf{Metric}} & \multicolumn{1}{c}{\textbf{RumourEval-19}} & \multicolumn{1}{c}{\textbf{Weibo-comp}} \\ 
        \hline
        Total News & 445 & 6,444 \\
        Real News & 138 & 2,863 \\
        Fake News & 185 & 3,581 \\
            Unverified News & 122 & -- \\
        Comments Count & 7,990 & 102,372 \\
        Avg. Comments per News & 18 & 16 \\
        \hline
    \end{tabular}
    \vspace{5pt}
    \caption{The statistics of RumourEval-19 and weibo-comp}
    \label{tab:dataset_stats}
\end{table}

\subsection{Baseline Methods}
To comprehensively evaluate the performance of SARC, we compare it with representative baseline models, which include: basic text learning models LSTM, Text-CNN, BERT, and HAN; social context-based models, namely HSA-BLSTM and dEFEND; sentiment-based models Dual-Emo and MTEFN; and models integrating the capabilities of Large Language Models (LLMs), including BERT w/G, dEFEND w/G, ARG, and ARG-D. Here w/G means the models are supplied with comment features generated by GenFEND \cite{nan2024let}.

\begin{itemize}
    \item \textbf{LSTM} \cite{hochreiter1997long}: A classic sequential model that processes word sequence features through average pooling of hidden states, followed by a fully connected layer for prediction.

    \item \textbf{Text-CNN} \cite{chen2015convolutional}: A CNN-based text classifier which uses convolution kernels of varying sizes to capture multi-granular semantic patterns for fake news detection.

    \item \textbf{BERT} \cite{devlin2019bert}: Transformer-based pretrained language model fine-tuned with the [CLS] token's hidden state as news representation.

    \item \textbf{BERT w/G}: It introduces the comment features generated by GenFEND \cite{nan2024let} to the BERT model, forming a joint input of news content features and generated comment features to enhance detection capability.
    
    \item \textbf{HAN} \cite{yang2016hierarchical}: Hierarchical attention network with bi-GRU layers modeling word-level and sentence-level features through paragraph segmentation.

    \item \textbf{HSA-BLSTM} \cite{guo2018rumor}: Hierarchical attention network extracting features from words, posts, and sub-events across news and social contexts.

    \item \textbf{dEFEND} \cite{shu2019defend}: Co-attentive framework combining news sentences and user comments through sentence-comment co-attention mechanism.

  \item \textbf{dEFEND w/G} \cite{shu2019defend}: It supplements the dEFEND model with the comment features generated by GenFEND \cite{nan2024let}, forming a joint input of news content features, actual comment features, and generated comment features.
    
    \item \textbf{Dual-Emo} \cite{zhang2021mining}: HSA-BLSTM enhanced with publisher-user bidirectional emotional features for emotion-aware detection.

    \item \textbf{MTEFN} \cite{choudhry2022emotion}: Multi-task framework jointly optimizing emotion classification and fake news detection.

    \item \textbf{ARG} \cite{hu2024bad}: ARG realizes the feature interaction between news text and multi-perspective analyses generated by LLMs through a dual cross-attention mechanism, dynamically screens and fuses useful information, and finally generates fake news predictions through aggregated features.
    
    \item \textbf{ARG-D} \cite{hu2024bad}: ARG-D is a distilled version of ARG, which reuses the news encoder and classifier parameters of ARG to realize fake news detection without querying LLMs.
\end{itemize}

\begin{table}[htbp]
\centering
\scalebox{0.9}{
    \begin{tabular}{l|cc}
        \hline
        \textbf{Parameter} & \textbf{Weibo-comp} & \textbf{RumourEval-19} \\
        \hline
        Embedding dimension & 300 & 300 \\
        News encoding hidden dim & 256 & 256 \\
        Comment encoding hidden dim & 128 & 128 \\
        Comment aggregation FC dim & 256 & 256 \\
        Comment emotion feature dim & 1 & 1 \\
        Number of clusters & 3 & 3 \\
        \hline
        Batch size & 8 & 8 \\
        Learning rate & 1e-3 & 5e-4 \\
        Optimizer & \multicolumn{2}{c}{Adam} \\
        Dropout rate & \multicolumn{2}{c}{0.5} \\
            Training epochs & \multicolumn{2}{c}{20} \\
        \hline
    \end{tabular}}
\vspace{5pt}
\caption{Hyperparameter setting}
\label{tab:Hyperparameter setting}
\end{table}
For the RumourEval-19 dataset, we follow the official competition metrics: Macro-F1 and RMSE. For Weibo-comp, we adopt Accuracy, Precision, Recall, and F1-score as evaluation metrics.

\subsection{Implementation Details}
On RumourEval-19 and Weibo-comp, the datasets are split into 7:1:2 and 6:2:2 ratios for training, validation, and testing, respectively. As shown in Table~\ref{tab:Hyperparameter setting}, the embedding dimension is fixed at 300, with hidden dimensions of 256 for the news encoder and 128 for the comment encoder. The comment aggregation FC dimension is 256, and each comment’s emotion feature is represented by a single scalar. The role clustering module uses 3 clusters for both datasets.

For the Weibo-comp dataset, we extract sentiment features using a fine-tuned BERT model and perform Chinese word segmentation with Jieba; the embedding layer is initialized using Chinese Word Vectors \cite{P18-2023, qiu2018revisiting}. For RumourEval-19, sentiment features are obtained via VaderSentiment \cite{hutto2014vader}, tokenization is carried out using NLTK, and the embedding layer is initialized with GloVe embeddings \cite{pennington2014glove, carlson2025new}.

Training uses a batch size of 8 and the Adam optimizer, with learning rates of $1\times10^{-3}$ for Weibo-comp and $5\times10^{-4}$ for RumourEval-19. A dropout rate of 0.5 is applied, and training proceeds for 20 epochs. These hyperparameters are selected based on preliminary experiments to balance performance and efficiency.

\subsection{Overall Results}

The experimental results are shown in Table~\ref{tab:Performance comparison}, where the best results are shown in bold while the second best ones are underlined. It is observed that SARC significantly outperforms other comparative methods in all classification metrics. On the RumourEval-19 dataset, the Macro-F1 of SARC is 0.357, which is significantly higher than that of the existing best-performing Dual-BLSTM (Macro-F1 = 0.334). The RMSE value of SARC is 0.761, which is significantly lower than that of all comparative models. This indicates that SARC can also perform well when the dataset is small. 

\begin{table}[htbp]
    \centering
    \scalebox{0.80}{
    \begin{tabular}{l|cc|cccc}
            \hline
        & \multicolumn{2}{|c|}{RumourEval-19} & \multicolumn{4}{c}{Weibo-comp} \\
        \cline{1-7}
        Method & Macro-F1 & RMSE & Accuracy & Precision & Recall & F1-score \\
        \hline
        LSTM & 0.254 & 1.186 & 0.816 & 0.826 & 0.849 & 0.837 \\
        Text-CNN & 0.267 & 1.042 & 0.829 & 0.809 & 0.906 & 0.855 \\
        HAN & 0.277 & 0.901 & 0.846 & 0.845 & 0.885 & 0.865 \\
        BERT & 0.294 & 0.988 & 0.884 & 0.885 & 0.910 & 0.897 \\
        BERT w/G & 0.327 & 0.855 & 0.872 & 0.875 & 0.881 & 0.878 \\
        HSA-BLSTM & 0.310 & 0.975 & 0.890 & 0.893 & 0.912 & 0.902 \\
        Dual-Emo & \underline{0.334} & 0.824 & 0.903 & \underline{0.921} & 0.902 & 0.911 \\
        dEFEND & 0.327 & 1.136 & 0.901 & 0.914 & 0.908 & 0.911 \\
        dEFEND w/G & 0.315 & \underline{0.816} & \underline{0.920} & 0.918 & \underline{0.922} & \underline{0.920} \\ 
        MTEFN & 0.321 & 0.879 & 0.835 & 0.880 & 0.815 & 0.846 \\
        ARG & 0.294 & 0.850 & 0.911 & 0.911 & 0.903 & 0.907 \\
        ARG-D & 0.287 & 0.868 & 0.872 & 0.881 & 0.868 & 0.874 \\
        SARC & \textbf{0.357} & \textbf{0.761} & \textbf{0.969} & \textbf{0.967} & \textbf{0.967} & \textbf{0.967} \\
            \hline
    \end{tabular}}
    \vspace{5pt}
        \caption{Performance comparison between SARC and baselines}
    \label{tab:Performance comparison}
\end{table}

For the Weibo-comp dataset, SARC also achieves the best performance. Even compared with the second best results, SARC still exhibits remarkable improvements of 4.9\%, 4.6\%, 4.5\%, and 4.7\% in terms of Accuracy, Precision, Recall, and F1-score, respectively. The accuracy improvement of SARC over the baseline models ranges from 4.9\% to 15.3\%, proving its outstanding performance in detecting real and fake news. To sum up, the combination of the semantic features of news content and user role features by SARC breaks the limitations brought by simple sentiment analysis and text modeling, and achieves better experimental results.

\subsection{Ablation Studies}
\begin{figure}[htbp]
    \centering
        \includegraphics[scale=0.3]{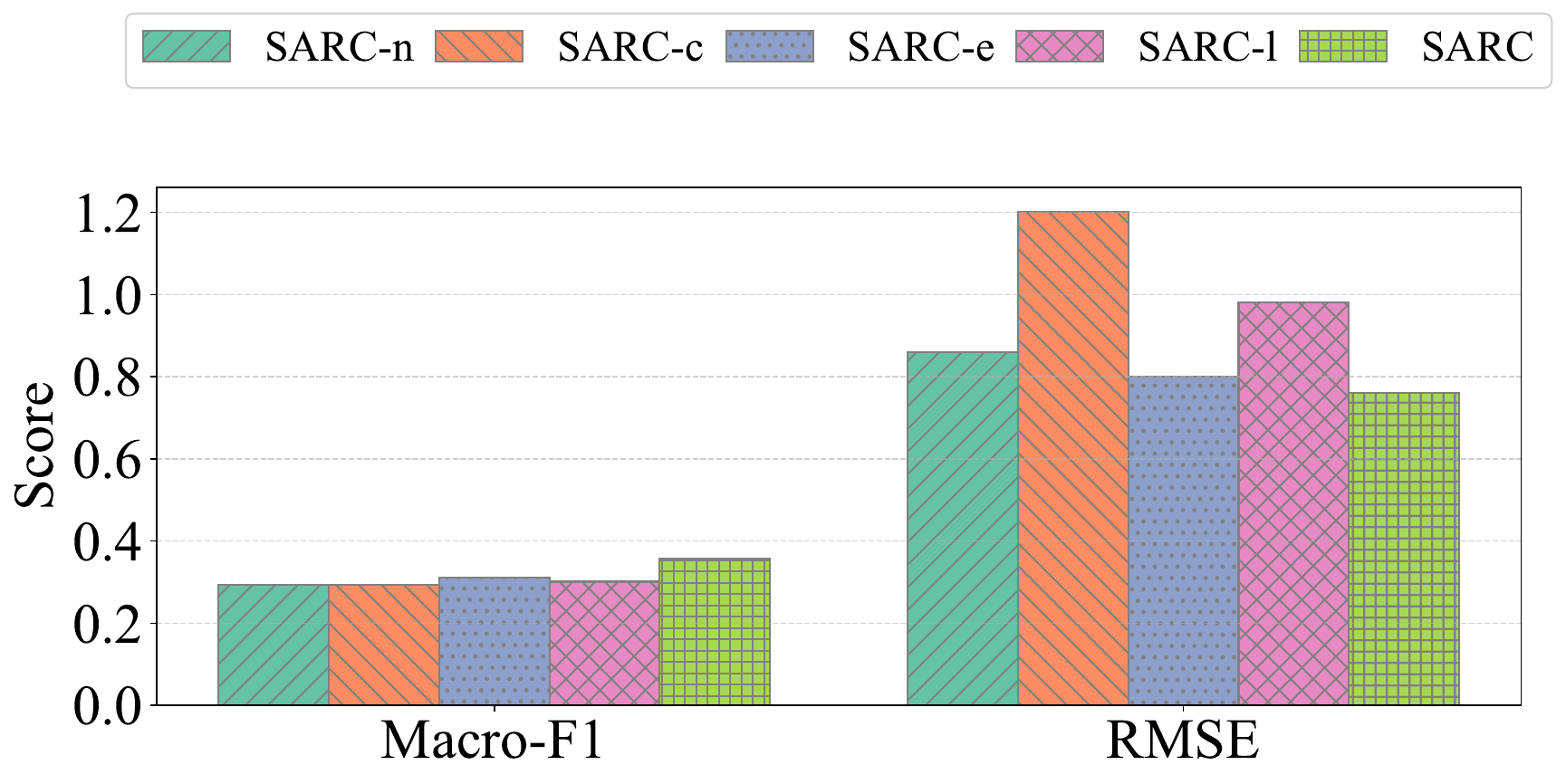}
    \caption{Performance comparison of SARC and its variant models on the RumourEval-19 dataset.}
    \Description{Performance comparison of SARC and its variant models on the RumourEval-19 dataset.}
    \label{ablation-rumou}
\end{figure}

\begin{figure}[htbp]
    \centering
    \includegraphics[scale=0.3]{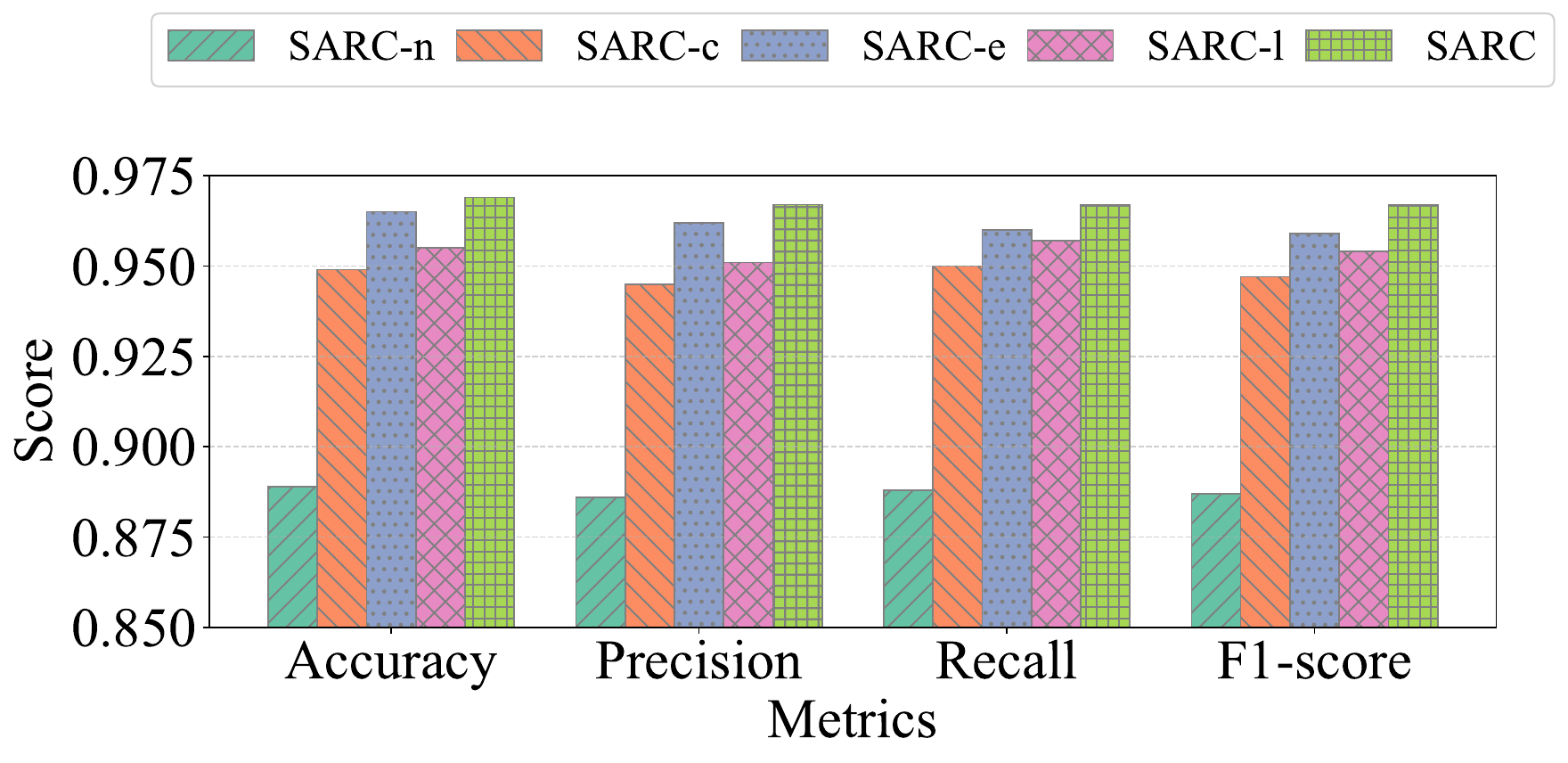}
    \caption{Performance comparison of SARC and its variant models on the Weibo-comp dataset.}
    \Description{Performance comparison of SARC and its variant models on the Weibo-comp dataset.}
    \label{ablation-weibo}
\end{figure}

We conduct four ablation studies to evaluate each module's contribution. First, SARC-n removes the news encoder, relying solely on the dynamic role clustering module. Second, SARC-c eliminates the dynamic role clustering module while retaining other components. Third, SARC-e discards comment sentiment features. Finally, SARC-l only uses the classification loss and discards the clustering loss. Results in Table~\ref{tab:Performance comparison} and visualizations in Figure~\ref{ablation-rumou} and Figure ~\ref{ablation-weibo} demonstrate each module's impact.

\begin{table}[htbp]
    \centering
    \scalebox{0.85}{
    \begin{tabular}{l|cc|cccc}
        \hline
        & \multicolumn{2}{|c|}{RumourEval-19} & \multicolumn{4}{c}{Weibo-comp} \\
        \cline{1-7}
        Method & Macro-F1 & RMSE & Accuracy & Precision & Recall & F1-score \\
        \hline
        SARC-n & 0.293 & 0.860 & 0.889 & 0.886 & 0.888 & 0.887 \\
        SARC-c & 0.293 & 1.201 & 0.949 & 0.945 & 0.950 & 0.947 \\
        SARC-e & 0.310 & 0.801 & 0.965 & 0.962 & 0.960 & 0.959 \\
        SARC-l & 0.301 & 0.981 & 0.955 & 0.951 & 0.957 & 0.954 \\
        SARC & \textbf{0.357} & \textbf{0.761} & \textbf{0.969} & \textbf{0.967} & \textbf{0.967} & \textbf{0.967} \\
        \hline
    \end{tabular}}
    \vspace{5pt}
    \caption{Ablation results of SARC }
    \label{tab:ablation}
\end{table}

Ablation results demonstrate the synergistic effects of SARC modules. As shown in Table~\ref{tab:ablation}, removing the news encoder (SARC-n) causes Macro-F1 on RumourEval-19 to plummet from 0.357 to 0.293, confirming this module's critical role in extracting core semantic features. Disabling the dynamic role clustering module (SARC-c) reduces Weibo-comp's F1-score by 1.3\% to 0.954, proving its effectiveness in enhancing detection through comment semantic clustering. Eliminating sentiment features (SARC-e) increases RMSE by 4\% on RumourEval-19, highlighting emotional signals' contribution. SARC-l, which only uses classification loss, also shows a decline in performance, indicating the synergy of joint optimization of classification and clustering. The progressive performance degradation (SARC-n < SARC-c < SARC-l < SARC-e < SARC) reveals module complementarity, with the full SARC achieving optimal metrics across datasets, thereby validating the combined efficacy of semantic modeling, clustering, and sentiment analysis.

\subsection{Hyperparameter Sensitivity Analysis}
\begin{figure}[htbp]
    \centering
        \includegraphics[scale=0.225]{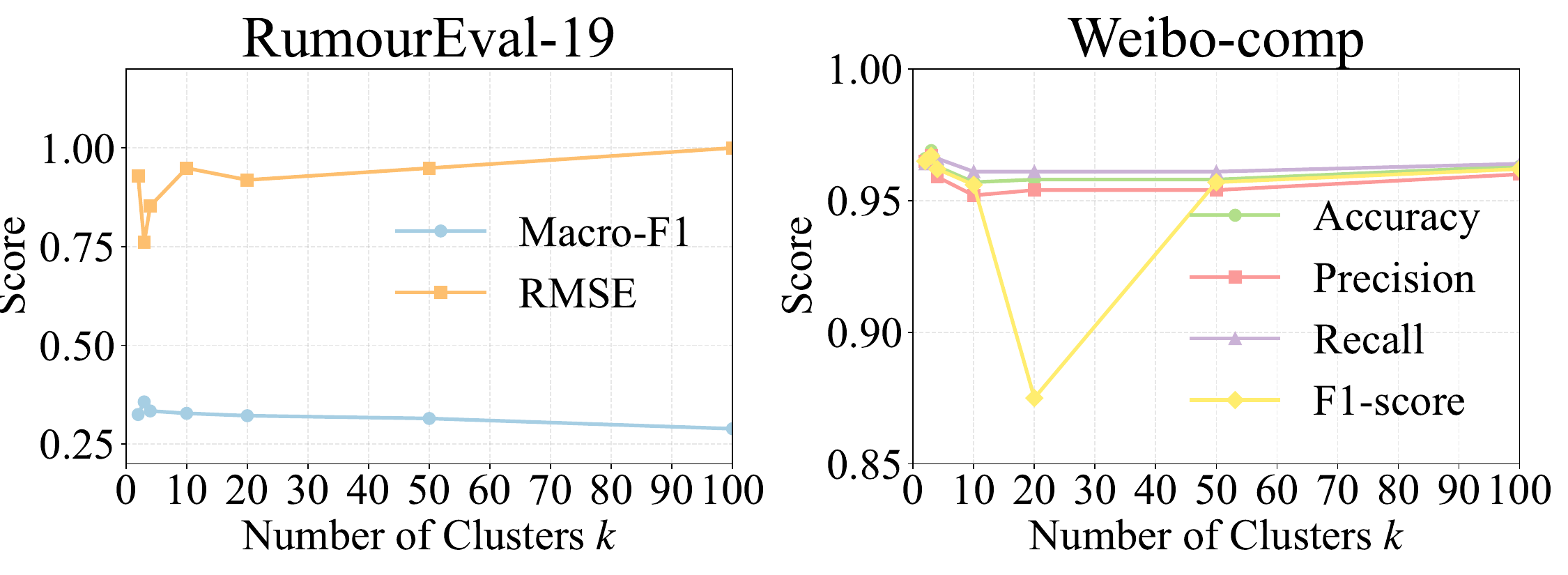}
    \caption{The impact of parameter $k$ on model performance.}
    \Description{The impact of parameter $k$ on model performance.}
    \label{fig:sensitivity_k_curve}
\end{figure}

We conducted two hyperparameter sensitivity analysis experiments, adjusting the number of clusters $k \in \{2, 3, 4, 5, 10, 20, 50, 100\}$ and the value of the clustering loss $\alpha \in \{0, 0.01, 0.05, 0.1, 0.2, 0.5, 1.0, 2.0\}$ in the loss function respectively. The experimental results are shown in Figure \ref{fig:sensitivity_k_curve}.

As can be seen from Figure \ref{fig:sensitivity_k_curve}, the number of clusters $k$ has a certain impact on the model performance, especially when $k$ lies within the range of $[1,10]$, and the model performs best when k=3. On RumourEval-19,  the performance of the model decreases as $k$ increases, but the degree of decrease is insignificant when $k$ becomes too large. On the Weibo dataset, the model's performance tends to be stable when $k\geq 10$. 

We conjecture this phenomenon is mainly due to the setting of primary and secondary weights of the loss function during training. Specifically, the clustering module is introduced as an auxiliary task in this model, and its loss is not the core target of training optimization. Though the value of $k$ changes, the classification loss still dominates, so the overall target optimization direction of the model will not shift drastically.

\begin{figure}[htbp]
\centering
\subfloat[k = 2]
{\label{fig:cluster-2}\includegraphics[width=0.48\linewidth]{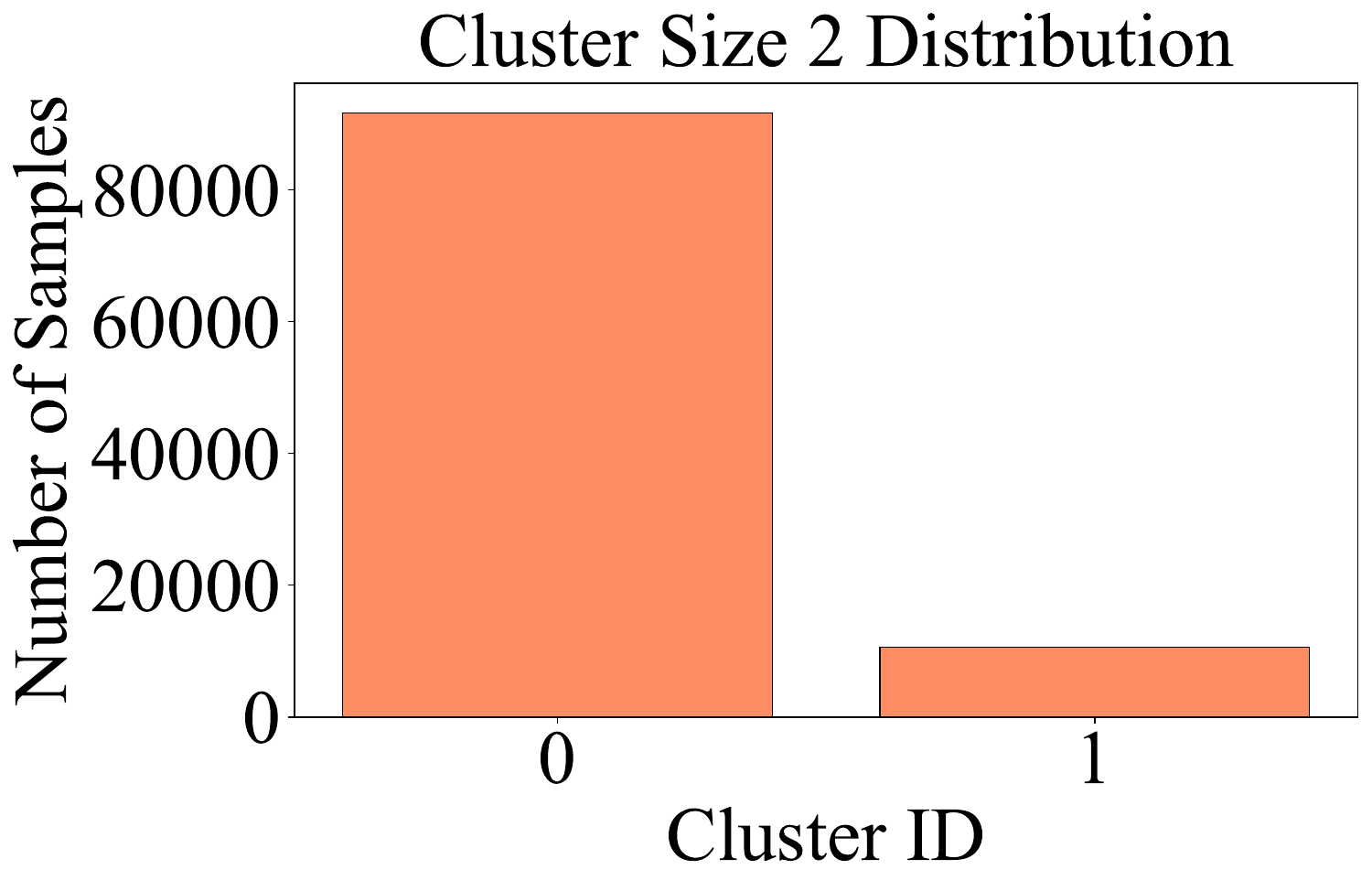}}
\hfill
\subfloat[k = 3]
{\label{fig:cluster-3}\includegraphics[width=0.48\linewidth]{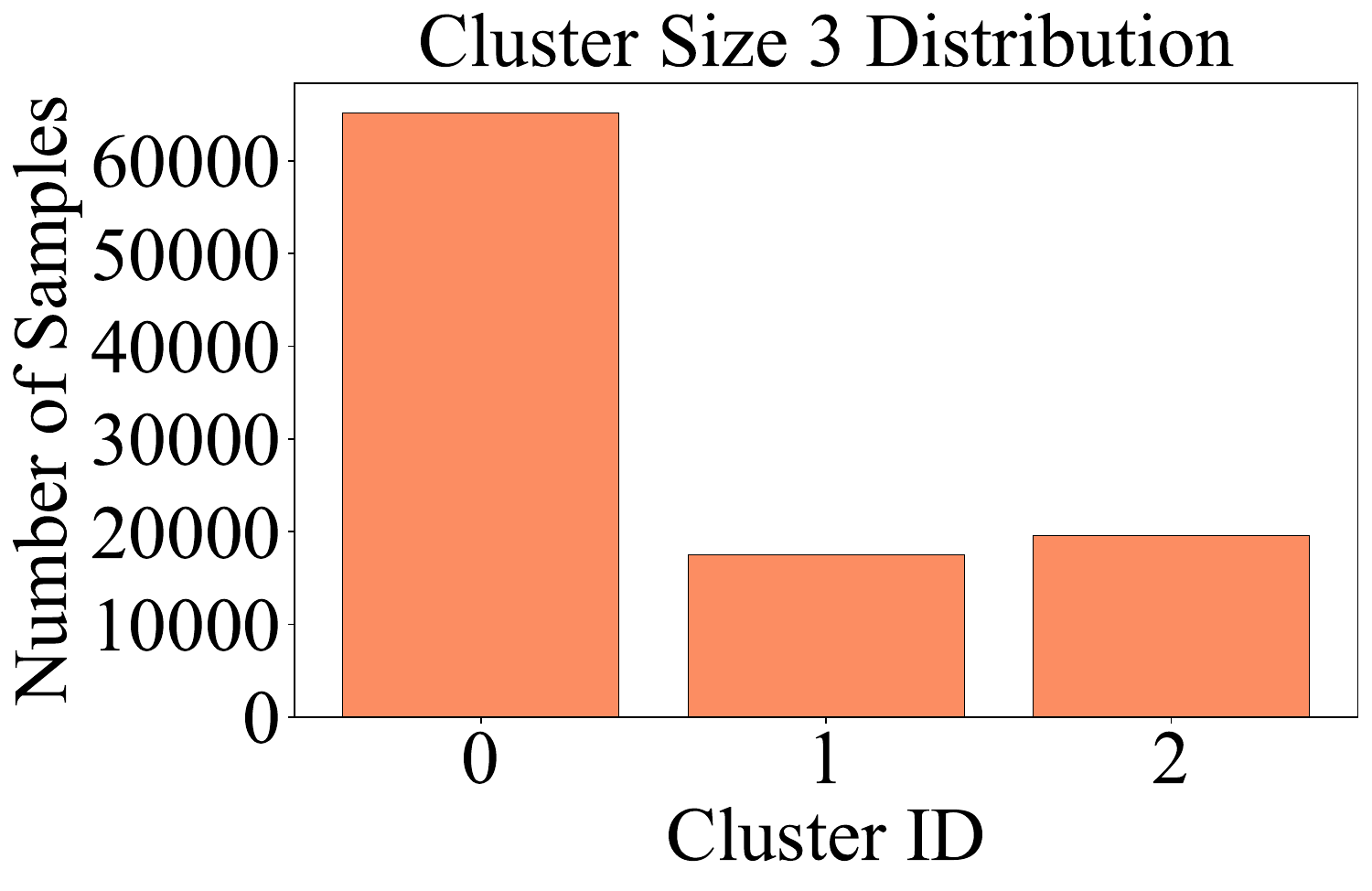}}
\vspace{-0.2cm}
\subfloat[k = 4]
{\label{fig:cluster-4}\includegraphics[width=0.48\linewidth]{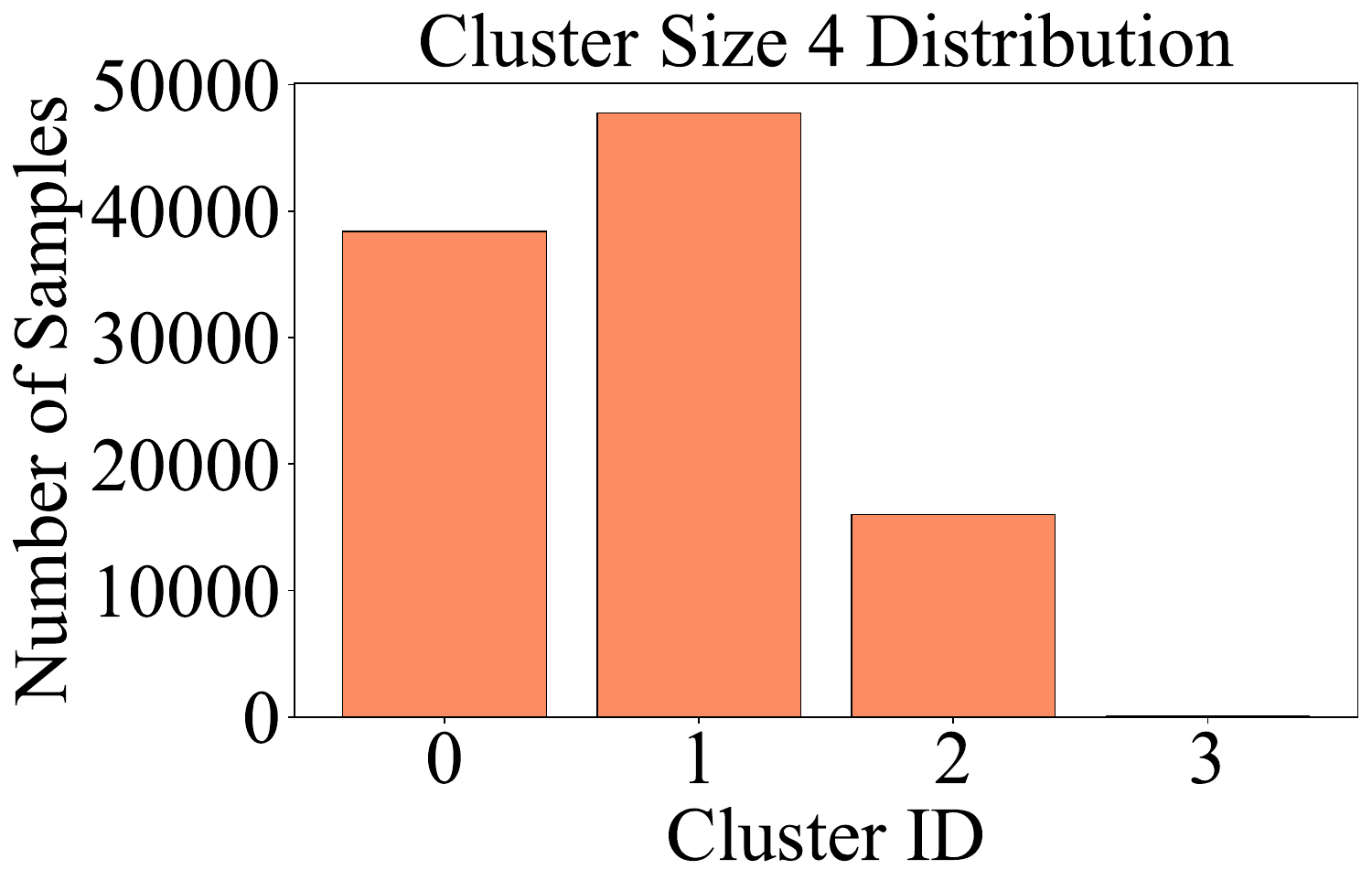}}
\hfill
\subfloat[k = 5]
{\label{fig:cluster-5}\includegraphics[width=0.48\linewidth]{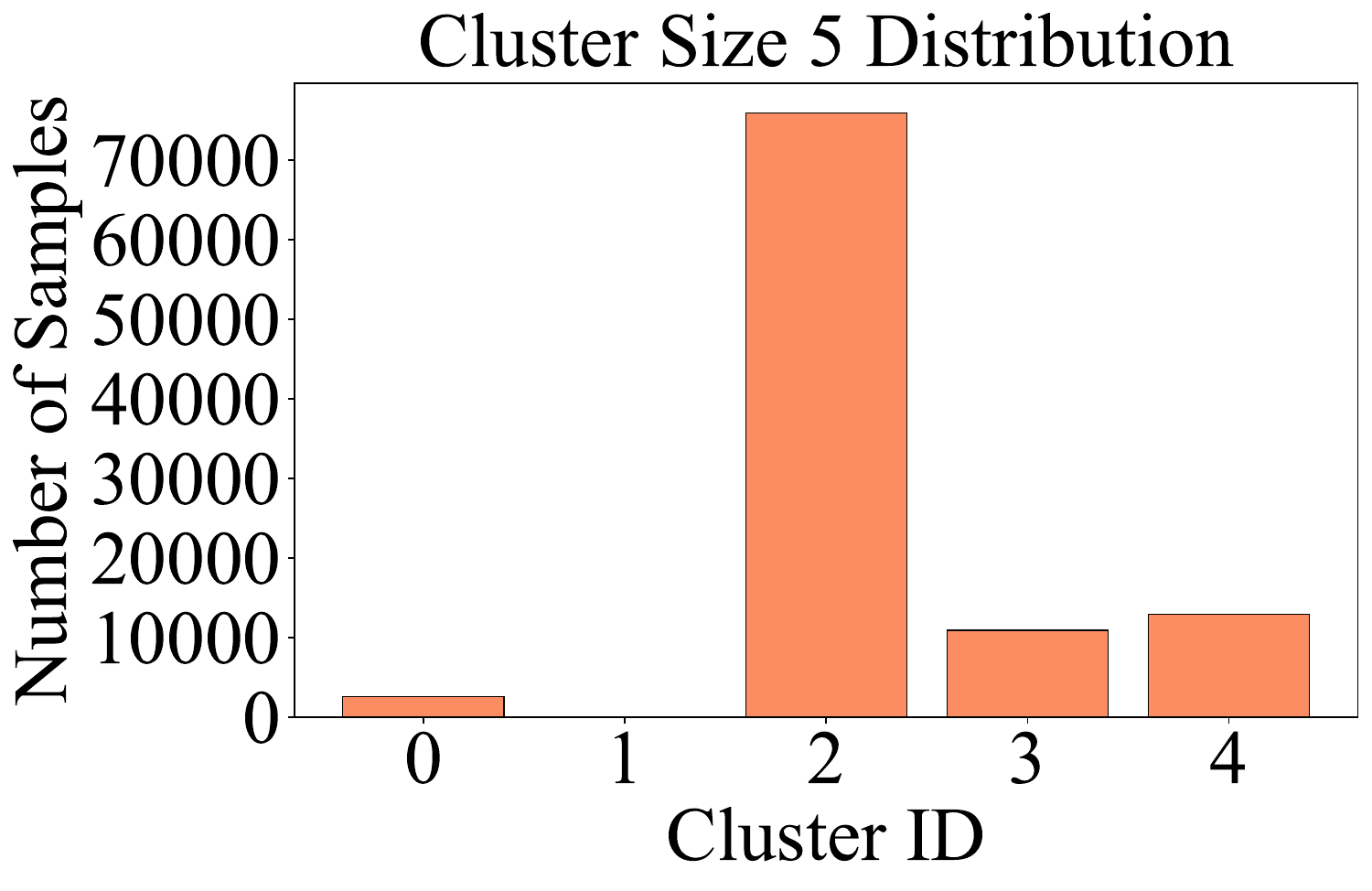}}
\vspace{-0.2cm}
\subfloat[k = 10]
{\label{fig:cluster-10}\includegraphics[width=0.48\linewidth]{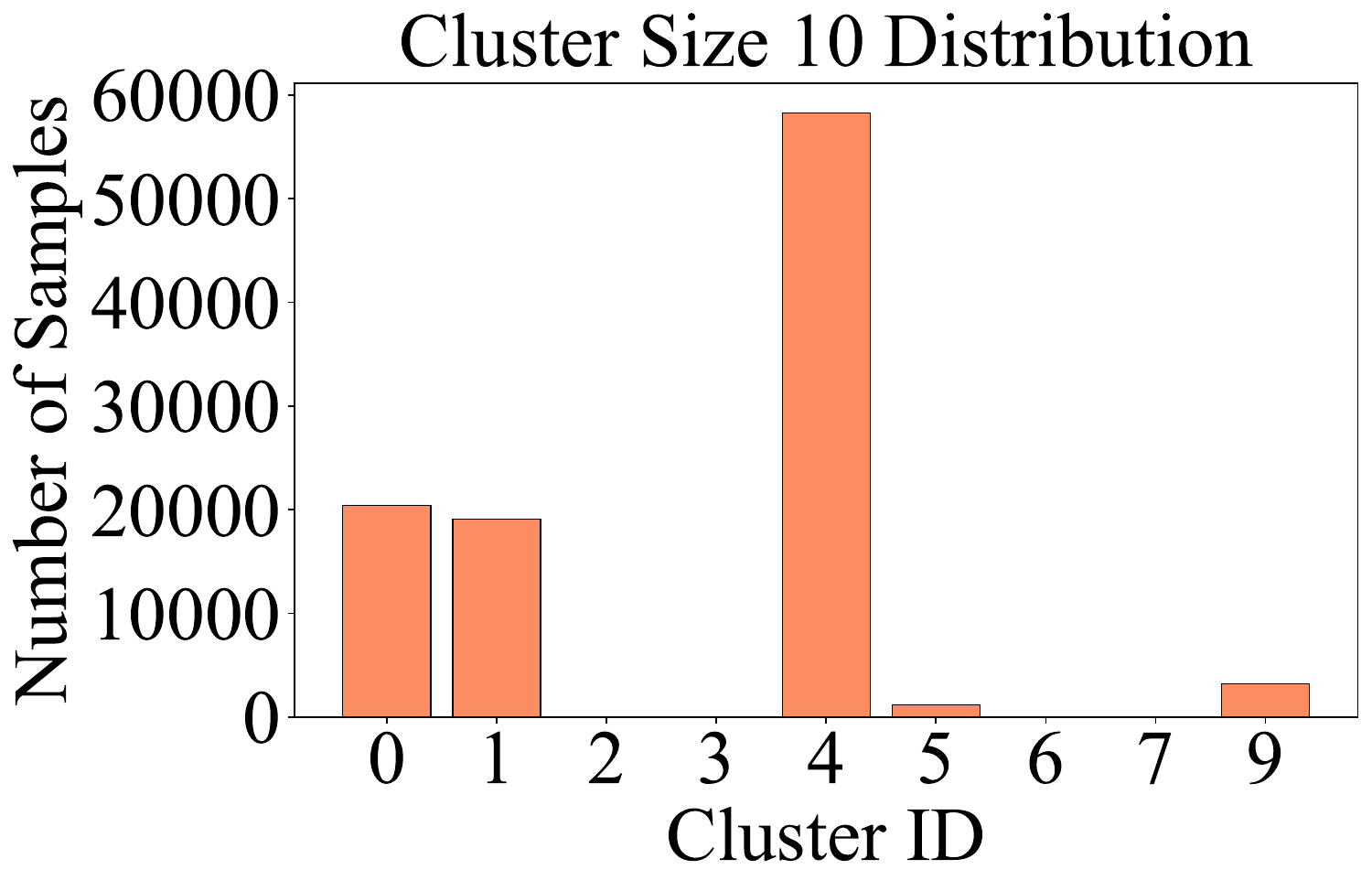}}
\hfill
\subfloat[k = 20]
{\label{fig:cluster-20}\includegraphics[width=0.48\linewidth]{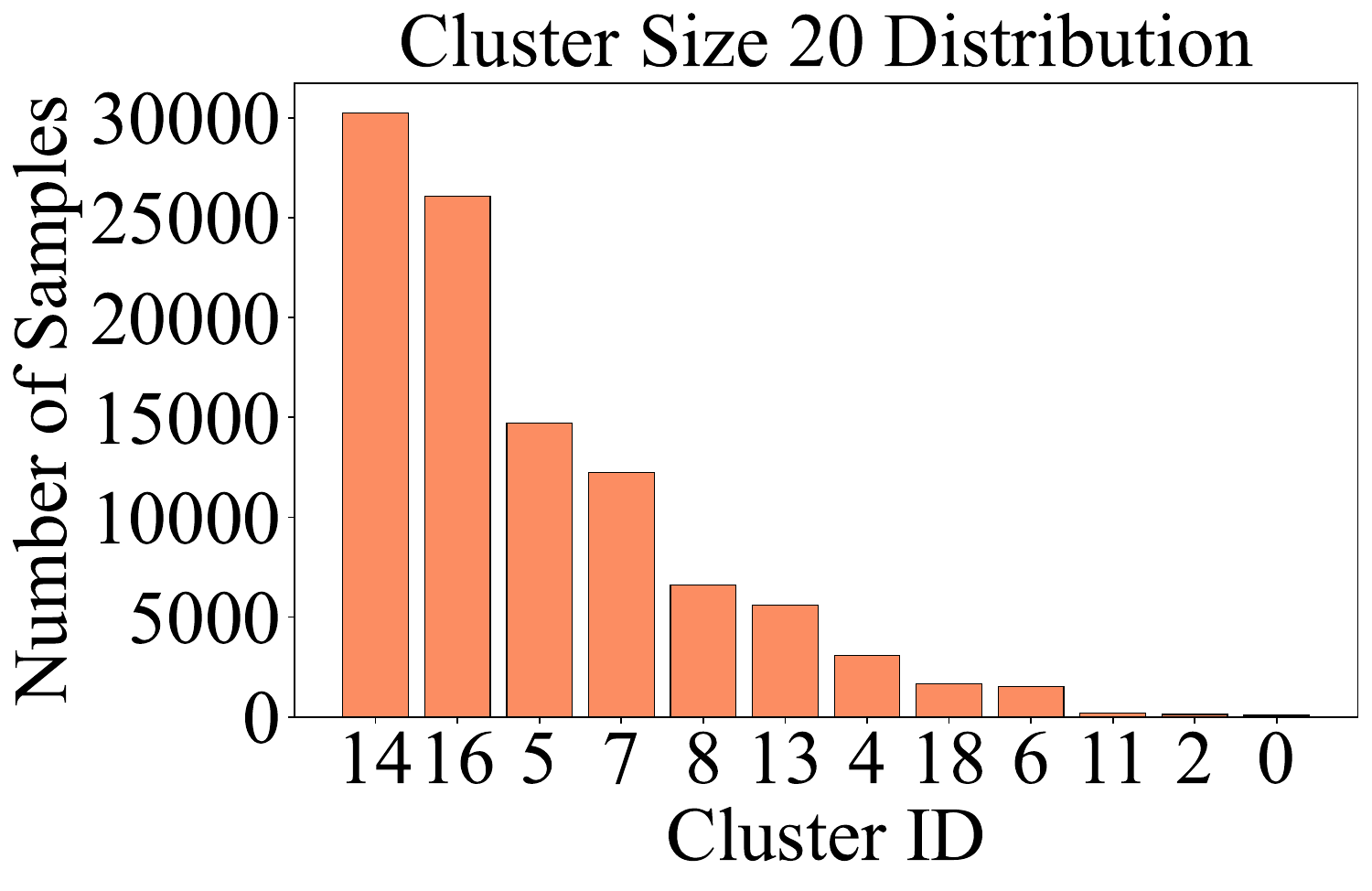}}
\vspace{-0.2cm}
\subfloat[k = 50]
{\label{fig:cluster-50}\includegraphics[width=0.48\linewidth]{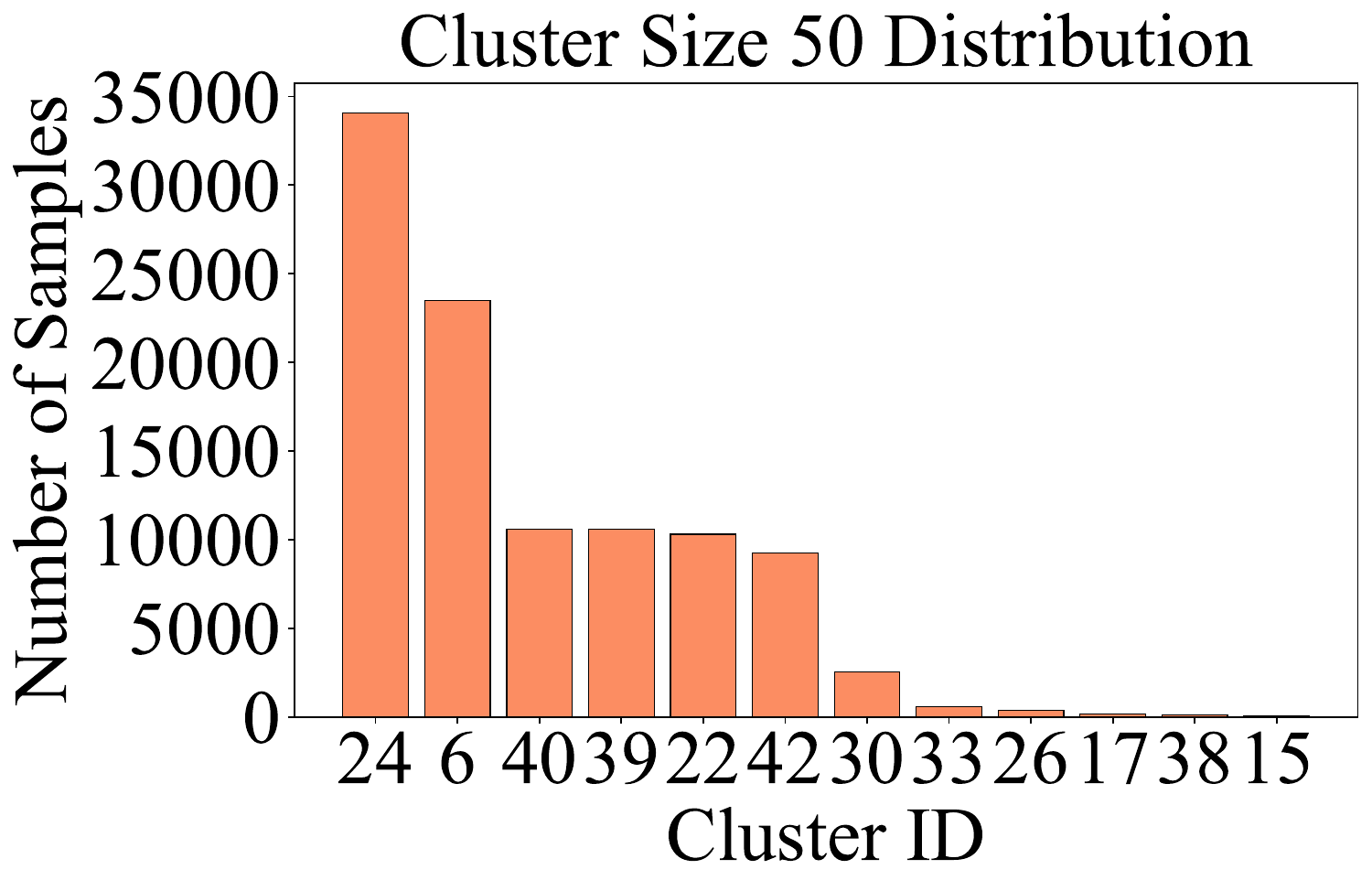}}
\hfill
\subfloat[k = 100]
{\label{fig:cluster-100}\includegraphics[width=0.48\linewidth]{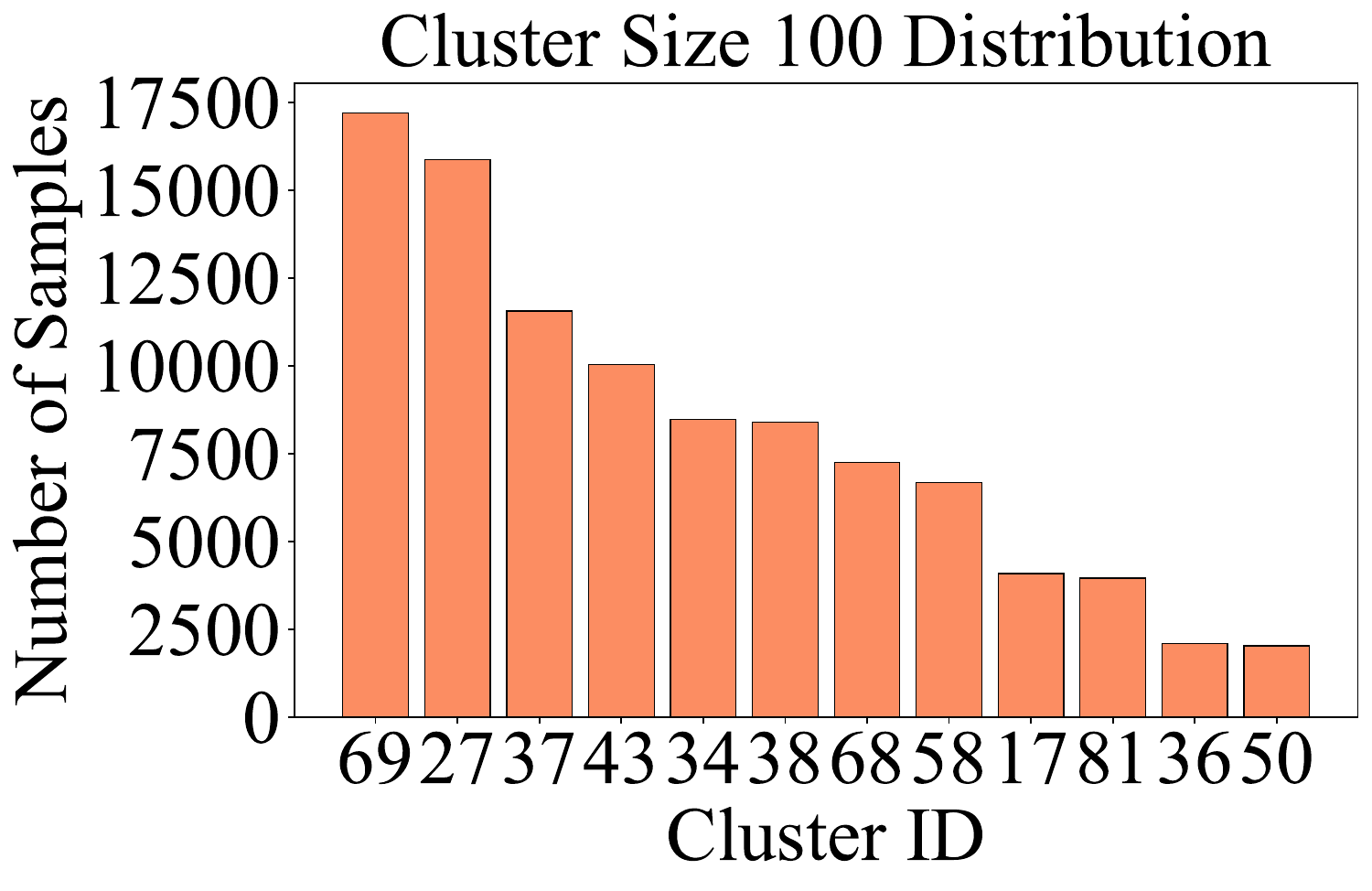}}
\caption{Visualization of clustering results with different numbers of clusters}
\label{fig:numClusters4x2}
\end{figure}

Moreover, because the clustering module adopts a soft assignment mechanism, the model can automatically adjust the correlation strength between samples and clusters, weakening the hard constraint of $k$. Consequently, the model possibly only activates a small number of effective clusters. As shown in Figure \ref{fig:numClusters4x2}, when the number of clusters is 4 or more, the sample size of some clusters is far lower than that of other clusters. This indicates that the model does not treat all clusters equally, but tends to focus on cluster structures with more concentrated information.

\begin{figure}[htbp]
    \centering
        \includegraphics[width=\linewidth]{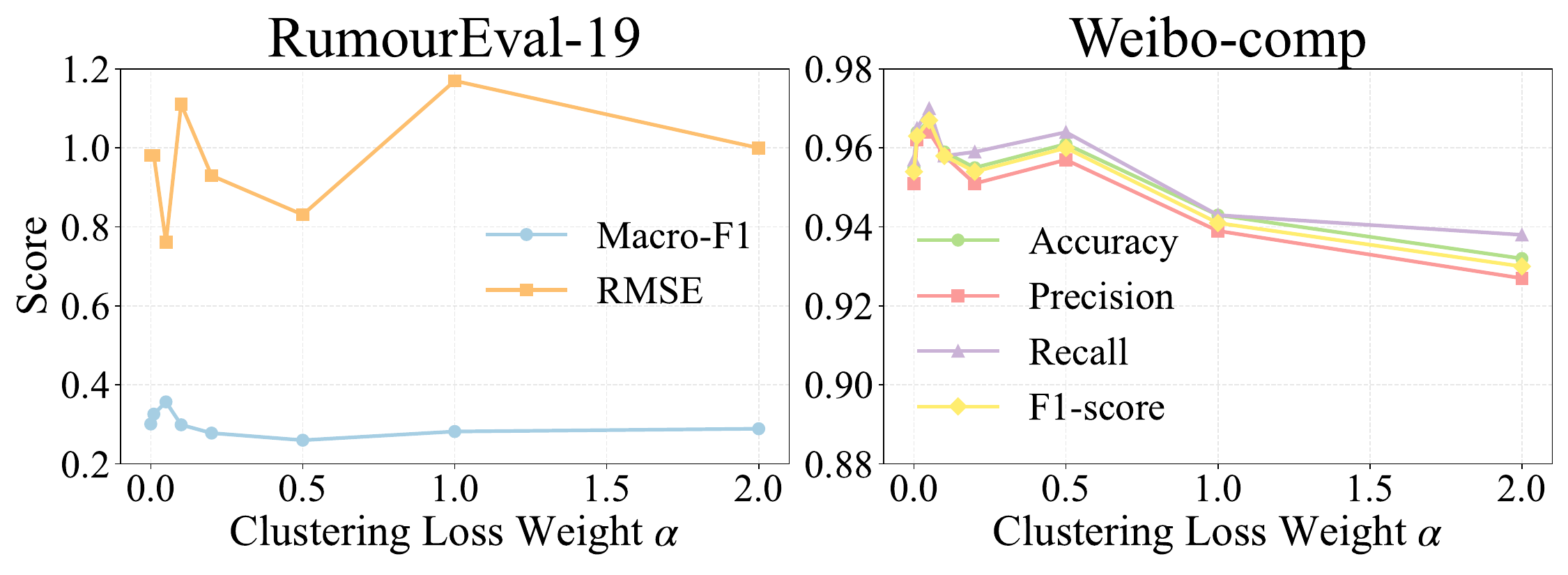}
    \caption{The impact of parameter $\alpha$ on model performance}
    \Description{The impact of parameter $\alpha$ on model performance}
    \label{fig:sensitivity_alpha_curve}
\end{figure}

Table~\ref{fig:sensitivity_alpha_curve} shows the model performance under different settings of the clustering loss weight $\alpha$. The experimental results indicate that the model is somewhat sensitive to this hyperparameter: an excessively small or large $\alpha$ will both lead to performance degradation, while a moderate range can significantly improve the model's detection performance. On RumourEval-19, the model performance generally declines as $\alpha$ increases and eventually plateaus. This behavior likely stems from the dataset’s small size: when the clustering-loss weight becomes too large, the model quickly bottoms out. Weibo, being much larger, exhibits a more linear sensitivity to $\alpha$, where the performance metrics keep sliding without any sign of leveling off as $\alpha$ continues to grow.

Specifically, when $\alpha = 0.05$, the model achieves optimal performance. On RumourEval-19, the Macro-F1 increases to 0.357 and the RMSE decreases to 0.761; on Weibo-comp, the Accuracy and F1-score reach 0.967 respectively, which is significantly better than other parameter settings. This suggests that under this weight, the clustering loss and classification loss achieve a well-balanced collaborative optimization.

When $\alpha$ is set too small (e.g., 0.01), the clustering module fails to provide sufficient guidance for the overall training, making it difficult to effectively exert its auxiliary function, resulting in only a slight improvement in classification performance. Conversely, when $\alpha$ takes a larger value (e.g., $\alpha\geq 0.1$), the clustering loss accounts for an excessively high proportion in training, which instead inhibits the classification learning process of the main task, and the model tends to suffer from the problem of optimization direction deviation. For example, when $\alpha = 2$, the Macro-F1 on RumourEval-19 drops to 0.289, and the Accuracy on Weibo-comp drops to 0.932, showing obvious performance degradation.

\subsection{Visual Analysis}

\begin{figure}[htbp]   
    \centering            
    \subfloat[fake news]
        {\label{fig:fake-news-pot}\includegraphics[width=0.45\linewidth] {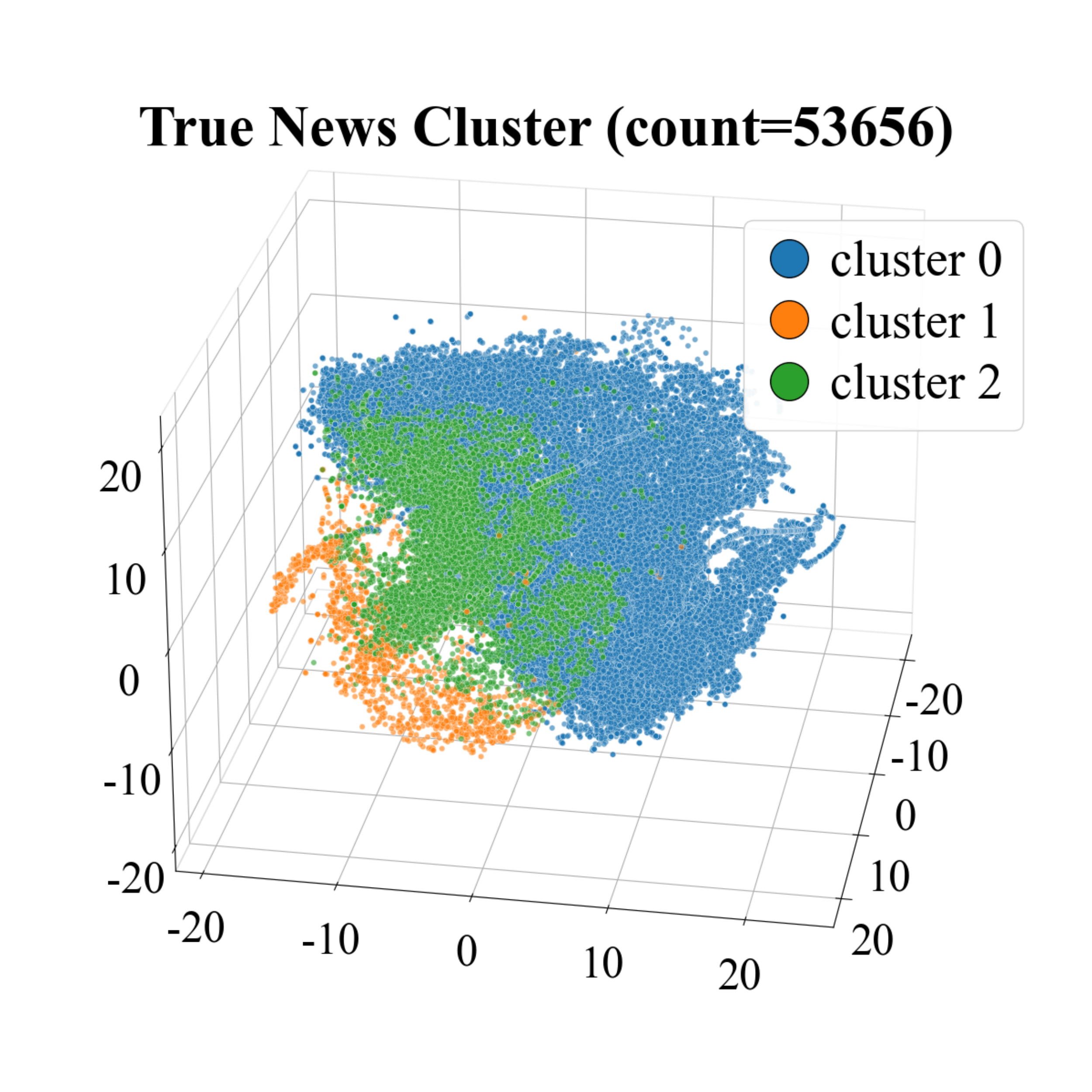}}
    \subfloat[true news]
        {\label{fig:true-news-pot}\includegraphics[width=0.45\linewidth] {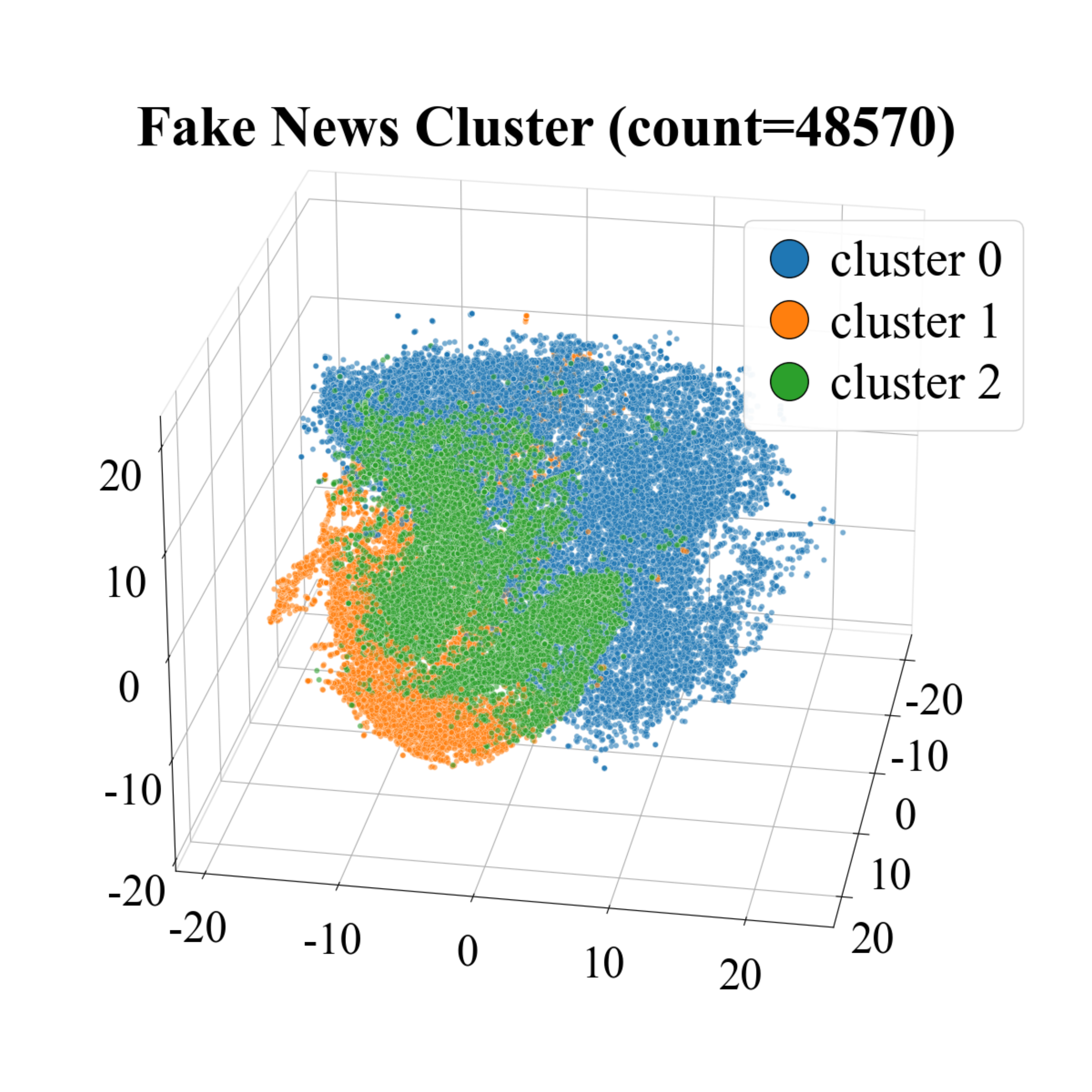}}
    \caption{Visualization results of features on Weibo-comp dataset} 
    \Description{Visualization results of features on Weibo-comp dataset}
    \label{fig:visualization}            
\end{figure}

Figure~\ref{fig:visualization} visualizes feature distributions of the Weibo dataset using t-SNE-based 3D projections with cluster-specific color coding, where Figure~\ref{fig:visualization}(a) displays fake news clusters (n=48,570), while Figure~\ref{fig:visualization}(b) shows true news distributions (n=53,656). Three clusters emerge: green (central), yellow (lower-left), and blue (upper-right), demonstrating clear cluster boundaries and intra-cluster compactness. Notably, fake news exhibits significantly more yellow points than true news, revealing comment content divergence and validating the dynamic clustering module's effectiveness.

\begin{figure}[htbp]
    \centering
    \includegraphics[width=0.8\linewidth]{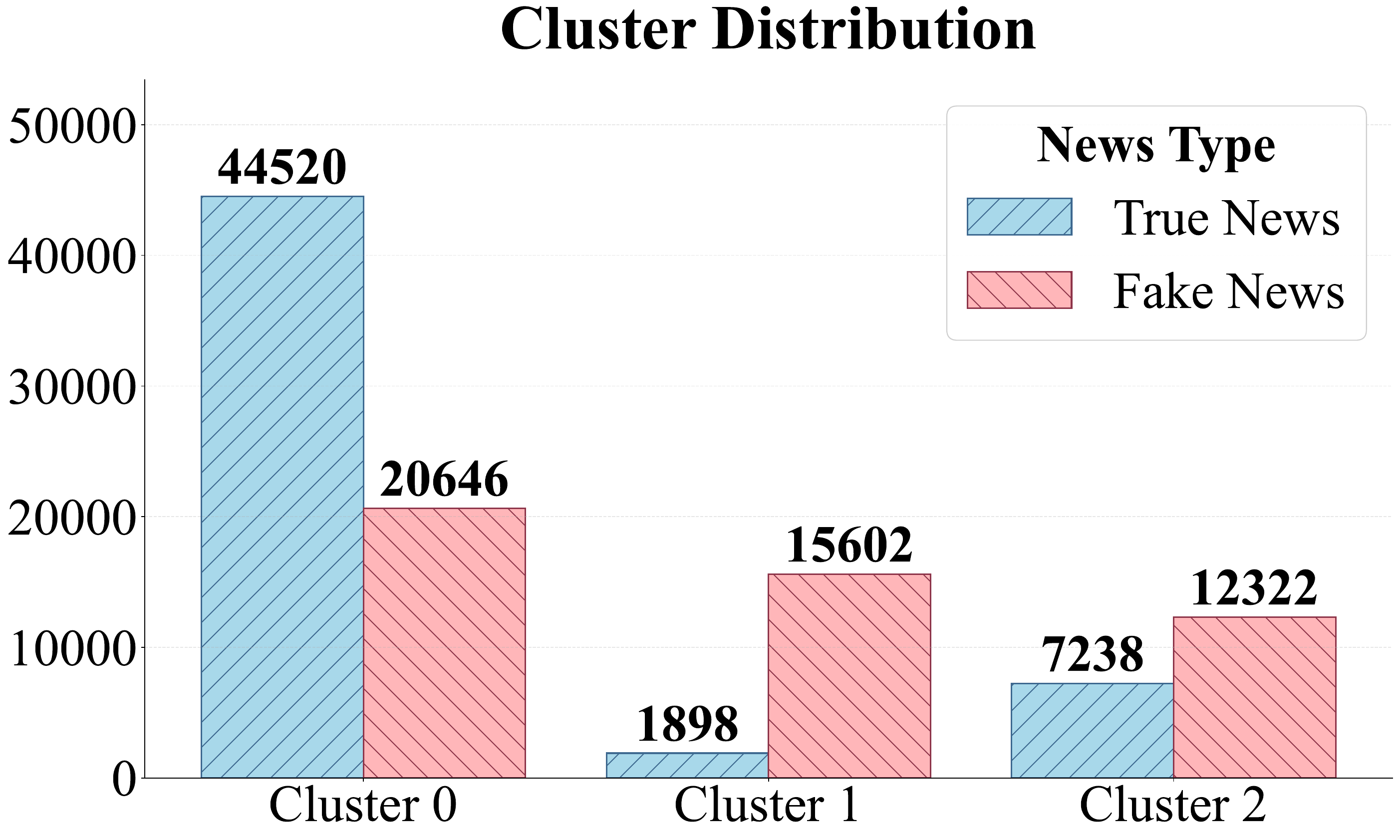}
    \caption{Cluster distribution on Weibo-comp dataset}
    \Description{Cluster distribution on Weibo-comp dataset}
    \label{fig:zhuzhuangtu}
\end{figure}

\begin{figure}[htbp]
\centering
\subfloat[cluster 0]
{\label{fig:word-cluster-0}\includegraphics[width=0.5\linewidth]{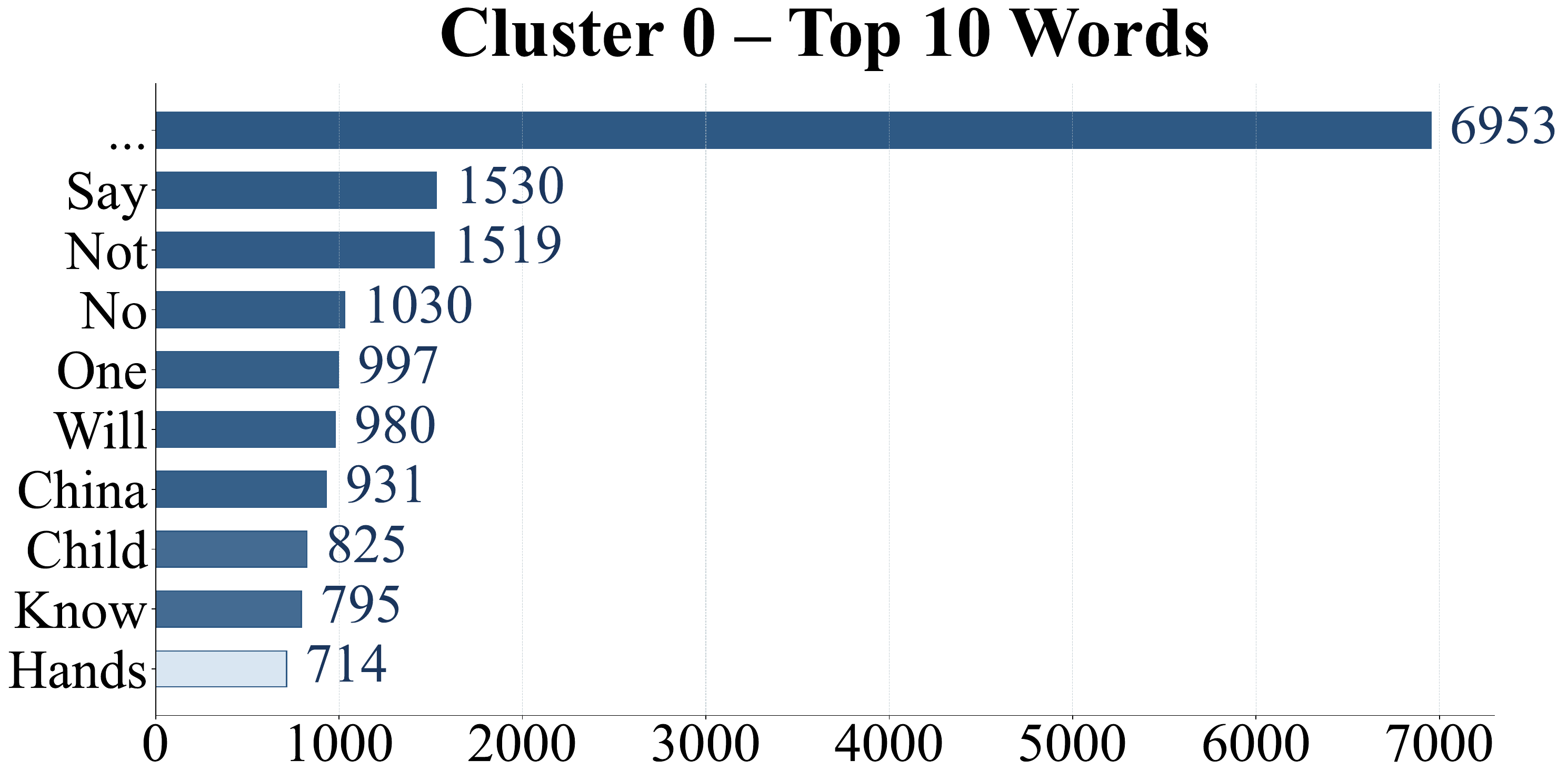}}
\hfill
\subfloat[cluster 2]
{\label{fig:word-cluster-2}\includegraphics[width=0.5\linewidth]{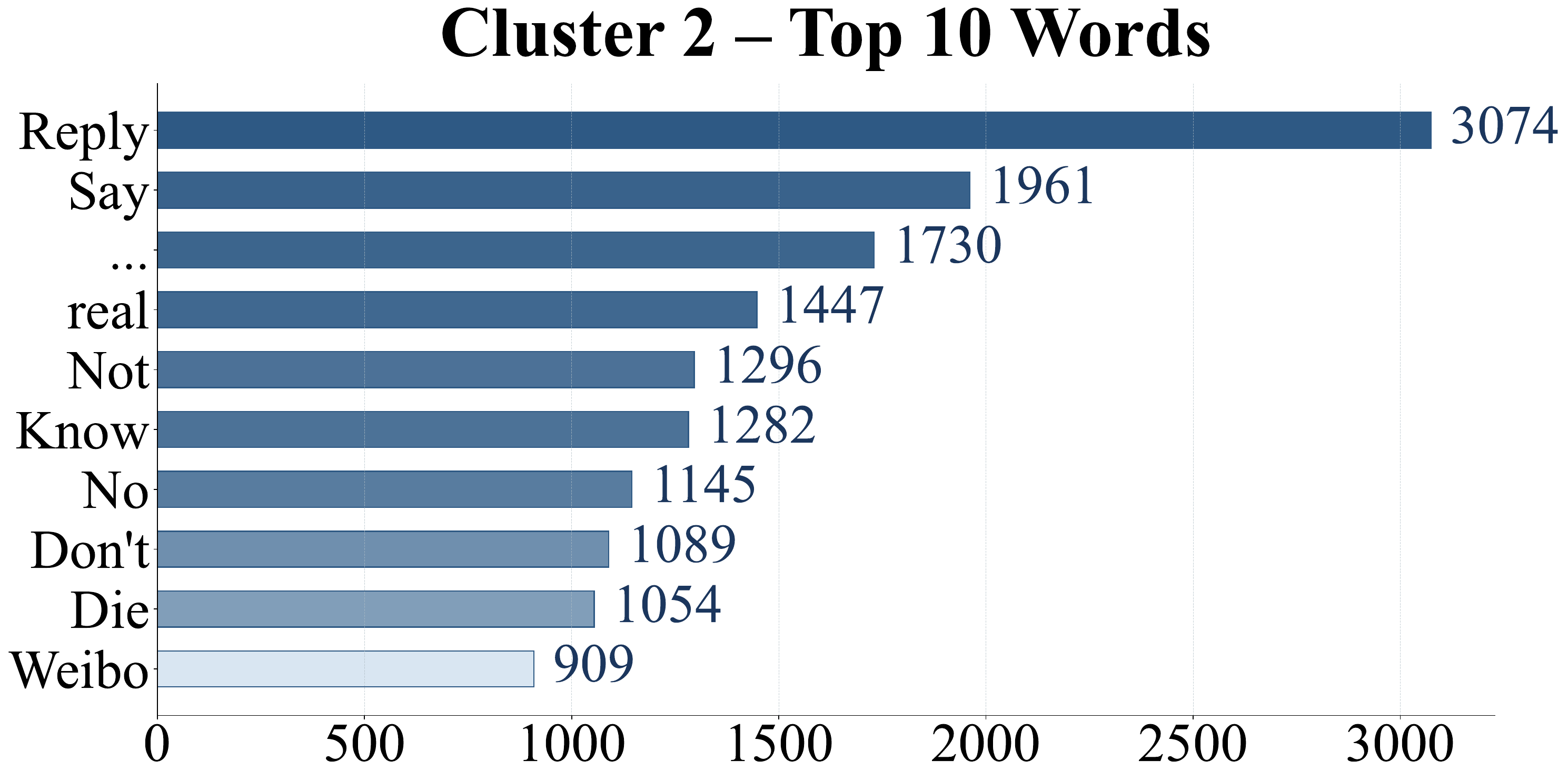}}
\vspace{0cm}
\subfloat[cluster 1]
{\label{fig:word-cluster-1}\includegraphics[width=0.7\linewidth]{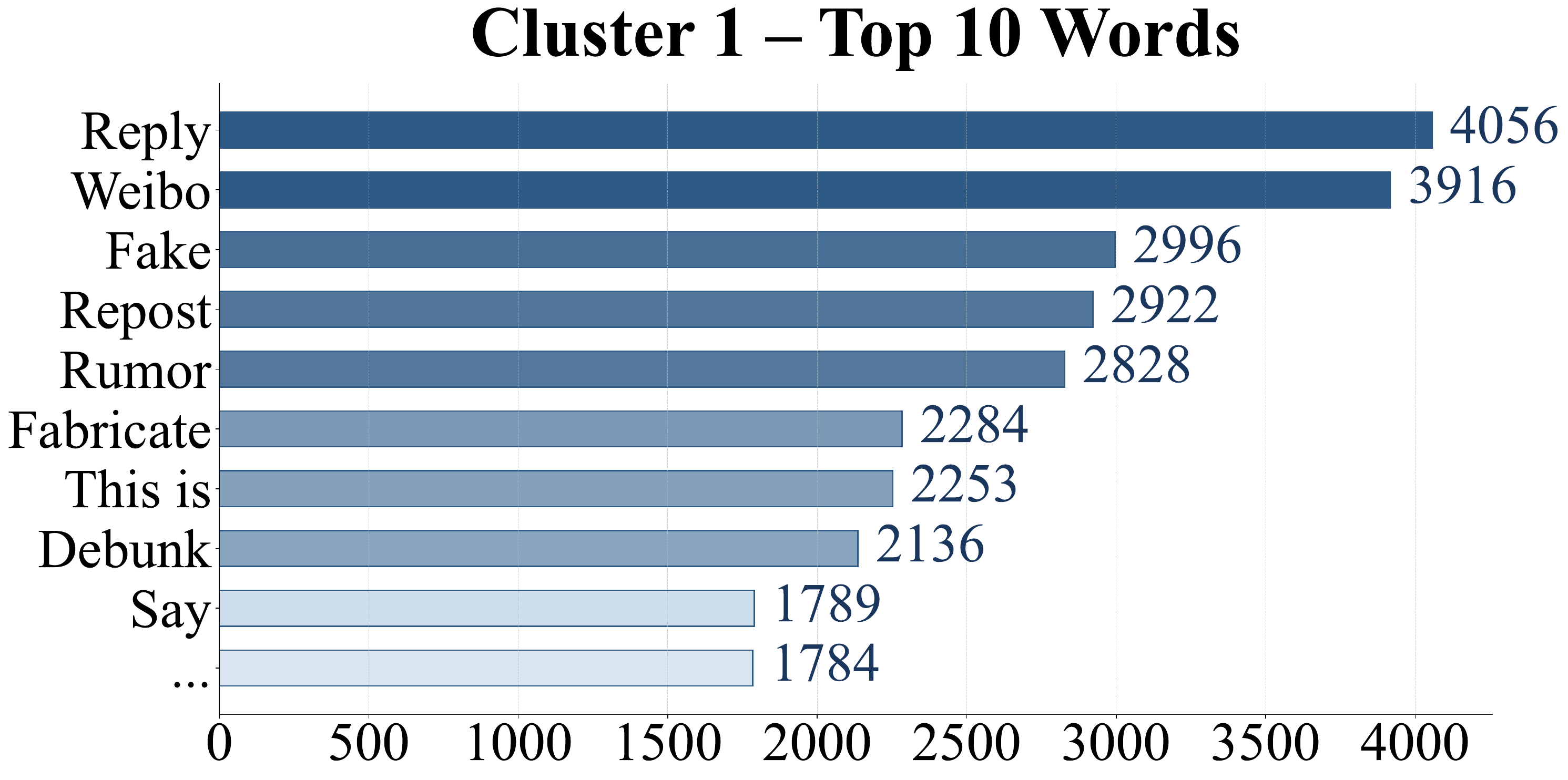}}
\caption{Word frequency statistics of clustering clusters}
\Description{Word frequency statistics of clustering clusters}
\label{fig:wordFrequency}
\end{figure}

\begin{figure}[htbp]
    \centering
    \includegraphics[width=1\linewidth]{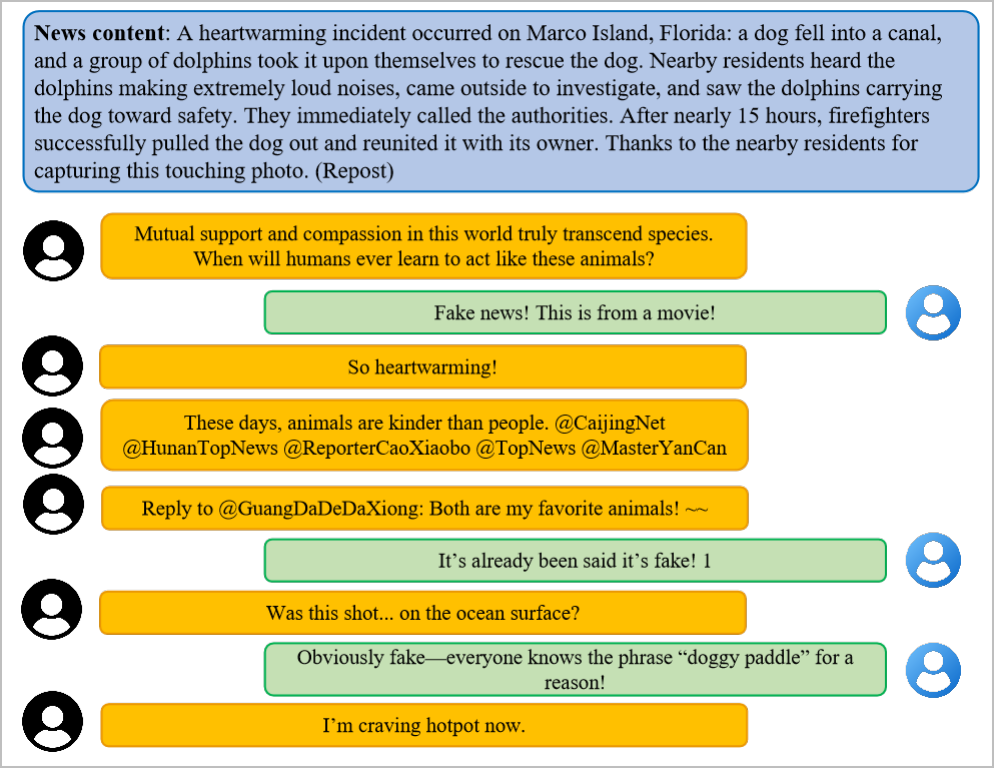}
    \caption{News example of cluster 1}
    \Description{news example of cluster 1}
    \label{fig:news_example_of_cluster_1}
\end{figure}

Figure~\ref{fig:zhuzhuangtu} quantifies cluster distribution differences. True news concentrates in Cluster 0 and Cluster 2 (Cluster 0 dominates with n=42,520, 86.05\%) reflecting semantic consistency. Fake news shows balanced inter-cluster distribution, with Cluster 1 comments being 8 times more prevalent. Word frequency analysis (Figure~\ref{fig:wordFrequency}) reveals Cluster 1's top terms, from which we see ``fake'', ``rumor'', ``Fabricate'' and ``Debunk'' ranked 3rd, 5th, 6th, and 8th respectively. Figure \ref{fig:news_example_of_cluster_1} shows a real-world fake news post and its comments. Cluster 1 (in green) mainly contains debunking and skeptical remarks. confirming Cluster 1 primarily hosts debunking comments in fake news. This is consistent with the result in Figure \ref{fig:zhuzhuangtu} that the number of Cluster 1 differs by 8 times, indicating a positive correlation between debunking behaviors and fake news.

\section{Conclusion}
In this paper, we tackled the fake news detection problem from a user role perspective using deep clustering. Learnable cluster centers dynamically group user comments, improving both behavior understanding and model explainability. Experiments on multiple datasets exhibited the model's outstanding performance, and ablation studies confirmed each module’s positive impact.

However, there remains room for improvement. First, the current clustering algorithm uses a fixed number of clusters, which may fail to adequately capture potentially important user roles in the data. Second, the BiGRU architecture combined with an attention mechanism may encounter performance bottlenecks when dealing with large-scale comment data. Future work could introduce deep clustering methods with adaptive cluster numbers and, when facing an overwhelming volume of comments, prioritize high-impact ones based on engagement metrics such as likes and shares to improve efficiency. Additionally, more effective strategies for fusing sentiment and textual features could be explored, along with the incorporation of multi-modal data. Furthermore, user-related metadata—such as account age and posting frequency—could be integrated into the role differentiation module to further enhance model performance.


\section{Ethical Considerations}
The proposed SARC framework, while aiming to improve fake news detection through sentiment-augmented role clustering, presents several ethical challenges. Biases in training data may propagate into the learned user role clusters and detection results, potentially reinforcing stereotypes or unfairly targeting specific groups. Since user comments often reflect cultural norms and linguistic nuances, a model trained in one context may misclassify in another, raising cross-cultural fairness concerns. Overly aggressive detection thresholds risk suppressing legitimate speech, especially from marginalized voices or those critical of authority, while false negatives could enable harmful misinformation to spread. The unsupervised nature of the clustering process may limit explainability, making it more difficult to identify and correct embedded biases. Moreover, such technology could be misused to silence dissent under the pretext of combating ``fake news''.

\bibliographystyle{ACM-Reference-Format}
\bibliography{reference}

\end{document}